\documentclass[10pt,twocolumn,letterpaper]{article}

\PassOptionsToPackage{table}{xcolor}
\usepackage[pagenumbers]{iccv}

\usepackage{cuted}
\usepackage{caption}
\usepackage{booktabs}
\usepackage{multirow}
\usepackage{amsmath,amssymb,amsfonts}
\usepackage{bm}
\usepackage{xspace}
\usepackage{graphicx}
\usepackage{enumitem}
\usepackage{colortbl}

\newcommand{\ArtTet}{ArtTet\xspace}
\definecolor{oursrow}{HTML}{DCEBFA}
\newcommand{\oc}{\cellcolor{oursrow}}

\usepackage{tikz}
\usepackage[skins]{tcolorbox}
\definecolor{cPhys}{HTML}{A6462E}
\definecolor{cDiv}{HTML}{6B7F2A}
\definecolor{cCon}{HTML}{2F5F8F}
\definecolor{cIdea}{HTML}{7A6A58}
\newtcolorbox{ideabox}[1]{enhanced,colback=#1!14!white,colframe=#1!80!black,boxrule=0.5pt,arc=2.5pt,left=4pt,right=4pt,top=2.5pt,bottom=2.5pt,boxsep=0pt,before skip=4pt plus 1pt,after skip=4pt plus 1pt,fontupper=\small}
\newcommand{\gapbox}[3]{\begin{ideabox}{#1}\textbf{#2} #3\end{ideabox}}
\newcommand{\mk}[2]{\tikz[baseline=(m.base)]{\node[circle,draw=#2!80!black,fill=#2!20!white,line width=0.45pt,inner sep=0pt,minimum size=1.3em,font=\sffamily\bfseries\scriptsize,text=#2!50!black](m){#1};}}
\newcommand{\mP}{\mk{P}{cPhys}\,}
\newcommand{\mD}{\mk{D}{cDiv}\,}
\newcommand{\mC}{\mk{C}{cCon}\,}
\definecolor{iccvblue}{rgb}{0.21,0.49,0.74}
\usepackage[pagebackref,breaklinks,colorlinks,allcolors=iccvblue]{hyperref}

\def\paperID{*****}
\def\confName{ICCV}
\def\confYear{2026}

\title{ArtTet: Watertight Part Meshes of Articulated Objects\\from Rigid-Body Rules and Partitioned Marching Tetrahedra}

\author{
Sylvia Yuan\textsuperscript{1,*} \quad
Dan Wang\textsuperscript{2,*} \quad
Ravi Ramamoorthi\textsuperscript{1} \quad
Xinrui Cui\textsuperscript{3,\dag}
\\[4pt]
\textsuperscript{1}University of California San Diego \quad
\textsuperscript{2}University of Arizona \quad
\textsuperscript{3}University of North Texas
\\[4pt]
{\small \textsuperscript{*}Equal contribution \quad \textsuperscript{\dag}Corresponding author}
}

\begin{document}
\maketitle
\begin{strip}
\centering
\includegraphics[width=\linewidth]{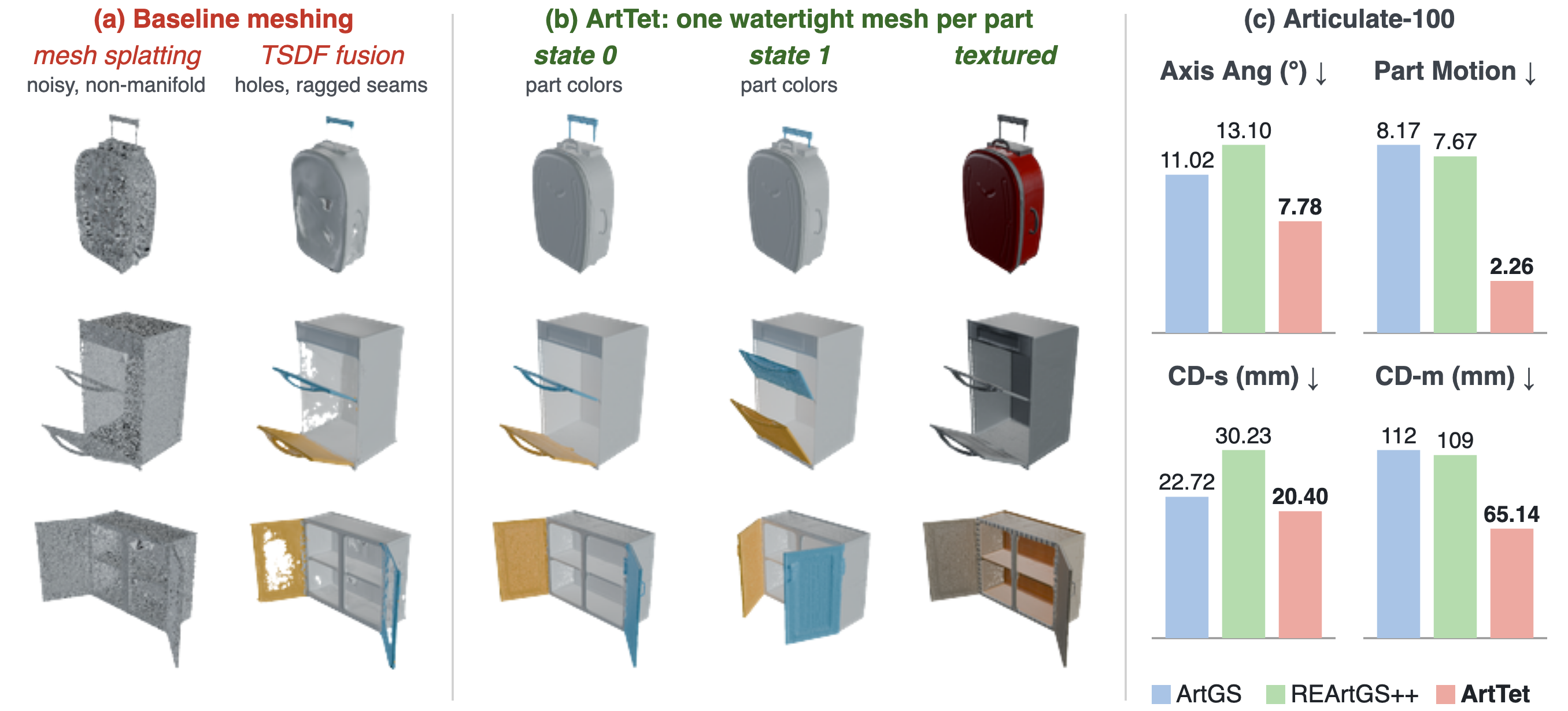}
\captionof{figure}{\textbf{\ArtTet reconstructs an articulated object as one watertight mesh per rigid part, with the joints that move them.} \textbf{(a)}~Baseline meshing of the same triangle reconstruction. Mesh splatting, in our adaptation of the restricted Delaunay step of MeshSplatting~\cite{meshsplatting}, gives a noisy, non-manifold global surface that seals openings, and TSDF fusion of the same reconstruction leaves holes and ragged seams between parts. \textbf{(b)}~\ArtTet divides the object into rigid parts with dynamic- and geometric-constrained articulation and ownership inference and conquers each part with its own exclusive signed distance field. Each object is shown colored by part in both observed states, and textured. \textbf{(c)}~Joint and geometry accuracy on Articulate-100.}
\label{fig:teaser}
\end{strip}

\begin{abstract}
We present \ArtTet, a method that reconstructs an articulated object from multi-view observations of two articulation states and outputs one watertight mesh per rigid part together with the joints that move them. Mesh splatting gives explicit triangles, but it treats the object as one scene. It extracts a single global surface and cannot keep the triangles of different parts apart; the mesh it exports is layered, self-intersecting, and full of holes. \ArtTet follows a divide-and-conquer strategy instead. We first divide the object into rigid parts with dynamic- and geometric-constrained articulation and ownership inference. A multibody registration of the two states recovers the parts and their screw joints. Rigid-body rules on attachment, contact, and swept motion then correct joint types and part labels without any part annotation. We then divide space into part cells with an articulated tetrahedral partition, and conquer each cell with its own differentiable signed distance field on a shared tetrahedral grid. In these partitioned marching tetrahedra the fields are mutually exclusive by construction, so the extracted part meshes are closed and never overlap. All part meshes are optimized jointly from the images of both states through the recovered joints. On Articulate-100, \ArtTet gives the most accurate axis directions, part motions, and part geometry over all objects among methods that use no part masks, and its part meshes have virtually no open boundary edges. We also show reconstructions of real captures from the SplArt dataset.
\end{abstract}

\section{Introduction}
\label{sec:intro}

Reconstructing articulated objects as explicit 3D assets is important for robotics, embodied AI, and interactive simulation. A cabinet replica, for example, requires separate surfaces for its frame, doors, and drawers, together with joints that reproduce their motion. Recovering such an asset requires consistent \textbf{part ownership, articulation, and surface geometry}: neighboring components must remain distinct, observations across states must describe the same rigid parts, and each reconstructed surface must define a coherent interior.

Recent articulated reconstruction methods~\cite{artgs,gaussianart,reartgspp,splart} estimate appearance and motion using implicit representations or Gaussian primitives~\cite{3dgs}. Their output is implicit or a cloud of primitives, so a mesh has to be extracted afterwards with TSDF fusion~\cite{tsdf} or marching cubes~\cite{marchingcubes}. Their conversion to meshes can introduce holes, merge neighboring components, or lose thin structures. Explicit mesh representations bring surface geometry into optimization~\cite{triangle-splatting,meshsplatting}, but connectivity alone does not resolve articulation: global meshing can connect adjacent rigid components, while independent meshing requires reliable part assignments. It leaves a central inference problem: \textbf{how can motion establish part ownership when masks are unavailable, and how can that ownership guide the reconstruction of closed part surfaces?}

This problem presents three coupled challenges.
First, motion is weak near joints. Points close to a hinge move little, so trajectory-based motion does not identify every part boundary reliably.
Second, ownership is uncertain at contacts. Where parts touch, repeated drawers can match the wrong component, and neighboring components compete for the same points.
Third, thin parts are observed incompletely. The back of a thin panel may remain hidden in both states, and a panel seen from one side admits multiple closed completions, some of which extend into neighboring components.

We present ArtTet, a method that reconstructs explicit part meshes and revolute or prismatic joints from posed multi-view RGB-D observations of two articulation states, without semantic part masks or joint annotations. \ArtTet follows a divide-and-conquer strategy in three stages that we mark throughout the paper (Fig.~\ref{fig:framework}): \mP dynamic- and geometric-constrained articulation and ownership inference, \mD the divide step, and \mC the conquer step.

\gapbox{cIdea}{Key idea.}{Divide and conquer. Rigid-body physics divides the \emph{object} into parts and joints without any annotation. An articulated tetrahedral partition then divides \emph{space} into one cell per part, and each cell is conquered by its own signed distance field whose interior excludes every other part. The extracted part meshes are therefore closed and never overlap.}

\mP We first divide the object into rigid parts. A multibody registration of the two observed point clouds recovers the parts and their screw joints. Rigid-body rules on attachment, contact, and swept motion then decide joint types and repair part labels while a triangle model is fitted to the images. \mD We then divide space. An articulated tetrahedral partition labels every cell of a Delaunay tetrahedralization as empty or as one of the parts, using free space seen in both states and the swept motion of each part. \mC Finally we conquer each part separately. Every part owns a signed distance field on one shared tetrahedral grid, and the fields are made mutually exclusive so that two parts can never claim the same point. Differentiable marching tetrahedra~\cite{dmtet} turns each field into a closed mesh, and all meshes are optimized together from the images of both states through the recovered joints (Fig.~\ref{fig:teaser}b). The exported meshes are the optimized meshes, apart from a minor cleanup that removes tiny disconnected pieces.

Our contributions are as follows.
\begin{itemize}[topsep=2pt,itemsep=1pt,parsep=0pt,leftmargin=*]
\item \mP\textbf{Dynamic- and geometric-constrained articulation and ownership inference (Sec.~\ref{sec:physics}).} Parts and joints come from a multibody registration of the two states. Rigid-body rules on hinge attachment, contact connectivity, and swept-motion collisions correct joint types and part labels with no masks and no joint annotation.
\item \mD\mC\textbf{Partitioned marching tetrahedra for articulated objects (Sec.~\ref{sec:meshing}).} An articulated tetrahedral partition divides space into part cells, and per-part differentiable signed distance fields with exclusive interiors conquer each cell. The result is one watertight, non-overlapping mesh per part, optimized end to end from two-state images without TSDF fusion or remeshing.
\end{itemize}
On Articulate-100~\cite{artmesh}, \ArtTet gives the lowest axis-angle, part-motion, and Chamfer errors over all objects among methods that do not use part masks, and its part meshes have virtually no open boundary edges (Fig.~\ref{fig:teaser}c). We also run \ArtTet on real captures from SplArt~\cite{splart} and score its joints against hand-labeled ones.

\section{Related Work}
\label{sec:related}

\paragraph{Articulated object reconstruction.}
Early methods model articulated objects with implicit fields. PARIS~\cite{paris} and Weng et al.~\cite{digitaltwinart} recover parts and motion from RGB or RGB-D captures of two states, Ditto~\cite{ditto} from point clouds, and category-level models~\cite{asdf,cla-nerf} learn articulation-conditioned fields. All represent shape implicitly, so a mesh must be extracted afterwards. Gaussian splatting made the representation explicit. ArtGS~\cite{artgs} initializes canonical Gaussians by matching two single-state reconstructions and assigns them to parts through learnable part centers, GaussianArt~\cite{gaussianart} blends a set of motion bases per Gaussian, and SplArt~\cite{splart}, ArticulatedGS~\cite{articulated-gs}, REArtGS~\cite{reartgs}, and REArtGS++~\cite{reartgspp} add part-aware or geometric constraints. All of them output primitives, and most extract their meshes by TSDF fusion of rendered depth. ArtMesh~\cite{artmesh} reconstructs triangle meshes directly and remeshes each part separately, but it relies on part masks to separate the parts. \ArtTet needs no part masks and produces closed part meshes without any post-hoc fusion.

\paragraph{Meshes from differentiable rendering.}
Surface reconstruction from splats usually converts the trained scene into a mesh afterwards~\cite{2dgs,gof,sugar}, and MILo~\cite{milo} extracts one at every iteration. Triangle Splatting~\cite{triangle-splatting} renders triangles directly but leaves them as an unconnected soup. MeshSplatting~\cite{meshsplatting} connects the soup with restricted Delaunay triangulation~\cite{delaunay} and exports an opaque mesh, yet it treats the input as one scene and keeps or drops each face independently, which folds layered soups into intersecting sheets. A separate line optimizes an implicit field whose zero level set is extracted differentiably. DMTet~\cite{dmtet} and FlexiCubes~\cite{flexicubes} do this on deformable grids, and the extracted mesh can be trained through a differentiable rasterizer~\cite{nvdiffrast,nvdiffrec}. These methods target static shapes. Labeling a Delaunay tetrahedralization by a minimum cut over visibility is a classic route to watertight surfaces~\cite{labatut,graphcut}. We extend both ideas to articulated objects. Free space is gathered through each joint, and several rigid parts share one grid, cannot overlap, and are supervised in two states.

\paragraph{Physical priors for articulation.}
Simulation enforces non-penetration with continuous collision detection and barrier potentials, as in Incremental Potential Contact~\cite{ipc}. Rigid registration~\cite{icp,umeyama} fits one rigid motion between two point sets, EM~\cite{em} fits several as a mixture, and screw theory~\cite{murray} gives each motion a joint interpretation. We combine these tools in a training-free registration, and we use contact and swept-motion collisions as evidence of part ownership rather than as a loss to minimize.

\section{Method}
\label{sec:method}

\begin{figure*}[t]
\centering
\includegraphics[width=\linewidth]{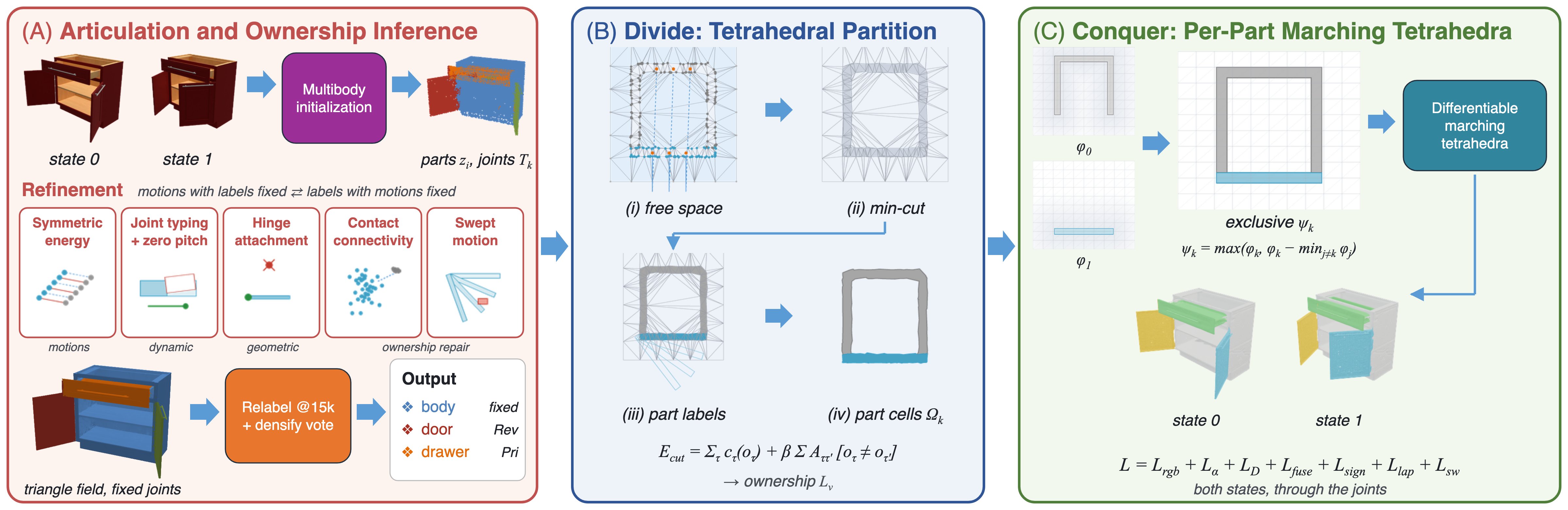}
\caption{\textbf{Overview of \ArtTet.} \textbf{(A)~Dynamic- and geometric-constrained articulation and ownership inference.} A multibody initialization splits the fused depth of both states into a static body and $K$ rigid parts with joints $\mathcal{T}_k$. A refinement alternates a symmetric energy on the motions with ownership repair on the labels, under dynamic and geometric joint constraints. A triangle model is then fitted to the images through the fixed joints. \textbf{(B)~Divide.} A Delaunay tetrahedralization of the observed points is labeled as empty or as one of the parts by a visibility minimum cut over both states and a swept-motion exclusion, giving one cell per part. \textbf{(C)~Conquer.} Each part owns a signed distance field $\phi_k$ on a shared tetrahedral grid. The exclusive fields $\psi_k$ cannot overlap, differentiable marching tetrahedra extracts one closed mesh per part, and all meshes are optimized from the images of both states through the joints.}
\label{fig:framework}
\end{figure*}
ArtTet addresses three coupled ambiguities: weak motion near joints, uncertain ownership at contacts, and incomplete observations of thin parts. We first infer rigid motion and repair ownership using geometric evidence, then aggregate motion-conditioned visibility on a tetrahedral graph, and finally optimize coupled part surfaces on a shared tetrahedral grid.

\paragraph{Problem Formulation.}
The input is $V$ posed RGB-D images $\{(I_v,D_v,A_v)\}_{v=1}^{V}$ of an articulated object in each of two states $s\in\{0,1\}$, where $A_v$ is the object silhouette, together with the number $K$ of moving parts. No part masks and no joint annotations are used. The output is one closed triangle mesh $\mathcal M_k$ per part in the canonical frame of state $0$, for $k=0,\ldots,K$ with $k=0$ the static body, and one revolute or prismatic joint per moving part.
We write the motion of part $k$ in pivot form,
\begin{equation}
\begin{gathered}
T_k^\eta(x)=R_k^\eta(x-p_k)+p_k+\eta d_k u_k,\\
R_k^\eta=\exp(\eta\theta_k[u_k]_\times),
\end{gathered}
\label{eq:arttet_motion}
\end{equation}
where $u_k$ is the unit joint axis, $p_k$ the pivot, taken as the point of the axis closest to the origin so that $u_k^\top p_k=0$, $\theta_k$ the rotation angle, $d_k$ the slide distance, and $\eta\in[0,1]$ the fraction of the motion between the two states.
A revolute joint has $d_k=0$; a prismatic joint has $\theta_k=0$ and its pivot is immaterial. We write $T_k^0=I$, $T_k^1=T_k$, and $T_0=I$.

\subsection{Dynamic- and Geometric-constrained Articulation and Ownership Inference}
\label{sec:arttet_registration}\label{sec:physics}
\mP This stage takes the depth of both states and $K$, and returns a part label for every observed point and one joint per moving part, typed as revolute or prismatic. We back-project and fuse the depth of each state into point sets $P^0$ and $P^1$ (Sec.~\ref{supp:impl}). A multibody initialization labels $P^0$ and fits one rigid motion per part. Two articulation constraints then decide each joint type, and a refinement improves the motions and the labels in turn (Fig.~\ref{fig:framework}A).

\paragraph{Multibody initialization.}
Each point $x_i\in P^0$ gets a label $z_i\in\{0,\ldots,K\}$, where label $0$ is the static body with the identity motion. Points of $P^0$ that are far from every point of $P^1$ are treated as moved. A grouping splits these moved points into $K$ clusters. Clusterings split the parts differently, so we form several groupings by spectral clustering, $k$-means, and clustering points that move together, each with two seeds (Sec.~\ref{supp:impl}). From each grouping we alternate three updates: every moved point takes the part whose motion brings it closest to $P^1$, every point $x_i$ of a part is paired with its nearest neighbor $y_i\in P^1$ under that part's motion, and every part's motion is refit to its pairs by
\begin{equation}
\min_{R\in SO(3),b\in\mathbb R^3}\sum_i\omega_i\|Rx_i+b-y_i\|_2^2,
\label{eq:arttet_procrustes}
\end{equation}
where $\omega_i\in\{0,1\}$ removes the $20\%$ of pairs with the largest distances. Eq.~\eqref{eq:arttet_procrustes} is solved by weighted centering and an SVD~\cite{umeyama}. Among all groupings, we keep the one whose final fit has the lowest residual. Its output is the labels $z_i$ and one rigid motion $(R_k,b_k)$ per moving part, which a registration in the reverse direction and an outlier cleanup complete (Sec.~\ref{supp:impl}).

\begin{figure}[t]
\centering
\includegraphics[width=\linewidth]{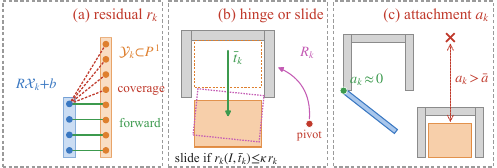}
\caption{\textbf{Joint typing.} \textbf{(a)}~$r_k$ adds the coverage of the target (red) to the forward fit (green). \textbf{(b)}~A drawer fit picks up a small rotation about a nearly orthogonal axis, but a pure translation explains it almost as well. \textbf{(c)}~A real hinge lies on its part, a spurious axis does not.}
\label{fig:typing}
\end{figure}

\paragraph{Dynamic articulation constraint.}
This constraint decides whether part $k$, with points $\mathcal{X}_k=\{x_i:z_i=k\}$ and motion $(R_k,b_k)$, is revolute or prismatic. A free rigid fit of a drawer often picks up a small rotation, whose axis is then wrong by almost $90^\circ$ (Fig.~\ref{fig:typing}b). We therefore score a motion $(R,b)$ of part $k$ by the symmetric residual (Fig.~\ref{fig:typing}a)
\begin{equation}
r_k(R,b)=\operatorname*{med}_{x\in\mathcal{X}_k} d\big(Rx+b,P^1\big)+\operatorname*{med}_{y\in\mathcal{Y}_k} d\big(y,R\mathcal{X}_k+b\big),
\label{eq:sy_residual}
\end{equation}
where $d(x,P)$ is the distance from $x$ to the nearest point of the set $P$, $\operatorname{med}$ is the median, and $\mathcal{Y}_k\subset P^1$ holds the points of state $1$ whose nearest point among all state-$0$ points, each moved by the motion of its own part, belongs to part $k$. The first term measures how well the moved part lands on $P^1$, and the second how well it covers the state-$1$ points it should explain. The joint is prismatic when the median displacement $\bar t_k$ between $\mathcal{X}_k$ and its mutual nearest neighbors in $P^1$ explains the part almost as well, $r_k(I,\bar t_k)\le\kappa\, r_k(R_k,b_k)$ with the slack factor $\kappa=1.3$, and revolute otherwise.

\paragraph{Geometric articulation constraint.}
A real hinge runs along the part it carries, so its axis line passes through or next to the part (Fig.~\ref{fig:typing}c). For a revolute joint with axis $u$ through the point $p$, we measure the distance from the part to its axis relative to the part size,
\begin{equation}
a_k(u,p)=\frac{1}{\operatorname{diag}(\mathcal{X}_k)}\min_{x\in\mathcal{X}_k}\big\|(I-uu^{\top})(x-p)\big\|,
\label{eq:sy_attach}
\end{equation}
where $\operatorname{diag}$ is the bounding-box diagonal. A revolute joint with a rotation angle below $\theta_s=13.5^\circ$ becomes prismatic in either of two cases: a translation $\hat t_k$ refit to the part explains it within the slack factor $\kappa'=1.1$, $r_k(I,\hat t_k)\le\kappa'\,r_k(R_k,b_k)$, or its axis lies away from the part, $a_k>\bar a$ with the threshold $\bar a=0.5$.

\paragraph{Refinement.}
The refinement starts from the labels and the typed joints of the initialization and improves the motions and the labels in turn. With the labels fixed, the motion $T_k$ of each moving part minimizes the symmetric energy
\begin{equation}
\begin{split}
E_k(T_k)={}&
\frac{\sum_{x\in\mathcal{X}_k}w_x\rho(d(T_kx,\mathcal{Y}_k)/\ell)}
     {\sum_{x\in\mathcal{X}_k}w_x+\epsilon}\\
&+
\frac{\sum_{y\in\mathcal{Y}_k}w_y\rho(d(T_k^{-1}y,\mathcal{X}_k)/\ell)}
     {\sum_{y\in\mathcal{Y}_k}w_y+\epsilon},
\end{split}
\label{eq:arttet_registration}
\end{equation}
with the Huber penalty $\rho$, the point spacing $\ell$, a small constant $\epsilon$, and a confidence $w\in[0,1]$ that is low for isolated points such as flying pixels (Sec.~\ref{supp:impl}). A prismatic part is refit as a pure translation, so the energy keeps every joint type.
With the motions fixed, an ownership repair reassigns points that registration mislabels, mostly near hinges and narrow gaps, where parts touch and points move little. It keeps every part one connected body that does not pass through the rest of the object along its joint (Sec.~\ref{supp:impl}).
We minimize the energy, repair ownership, minimize the energy again, and repeat the prismatic test of the geometric constraint with $r_k$, the second geometric pass. A part that still rotates by less than $5^\circ$ becomes prismatic, with its translation refit to $P^1$. The joints are fixed from here on, and we mark the fixed motions with a calligraphic letter, $\mathcal{T}_k^{\eta}$, with $\mathcal{T}_k=\mathcal{T}_k^{1}$.

\subsection{Triangle Field with Part Labels}
\label{sec:triangle}
This stage takes the images of both states, the joints, and the labeled points of $P^0$, and returns a canonical triangle model whose vertices carry part labels. The partition of Sec.~\ref{sec:meshing} uses these labels to assign the depth of state $1$ to parts. Following MeshSplatting~\cite{meshsplatting}, each vertex $\mathbf{q}_j$ carries a color, an opacity, and part probabilities $\pi_j$ over the $K+1$ parts, with the label $z_j=\arg\max_k\pi_{jk}$. The model is initialized from the labeled points. State $0$ is rendered as is, and state $1$ by moving every vertex with the joint of its part, $\mathcal{T}_{z_j}(\mathbf{q}_j)$. All vertex attributes are optimized on the images of both states with
\begin{equation}
\mathcal{L}_{\mathrm{tri}}=\mathcal{L}_{\mathrm{MS}}+\lambda_\pi\mathcal{L}_\pi,\qquad \mathcal{L}_\pi=-\frac{1}{J}\sum_{j=1}^{J}\log\pi_{jz_j},
\label{eq:tri}
\end{equation}
where $\mathcal{L}_{\mathrm{MS}}$ holds the photometric, depth, and normal losses of MeshSplatting, $J$ is the number of vertices, and $\mathcal{L}_\pi$ keeps each vertex on its current label. The joints stay fixed. The rigid-body rules of Sec.~\ref{sec:physics} also relabel the vertices during optimization (Sec.~\ref{supp:impl}).

\begin{figure}[t]
\centering
\includegraphics[width=\linewidth]{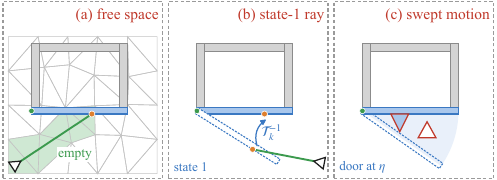}
\caption{\textbf{Articulated tetrahedral partition.} \textbf{(a)}~The segment from a camera to its observed depth crosses only empty space, so every tetrahedron on it gains a free-space vote. \textbf{(b)}~A ray of state $1$ that sees the door is moved into the canonical frame by $\mathcal{T}_k^{-1}$ of the door. \textbf{(c)}~Posed at $\eta$, the door sweeps the shaded region and passes through two tetrahedra labeled static (red outline). The one that touches the door goes to the door (blue), and the other goes to the outside (white).}
\label{fig:partition}
\end{figure}

\begin{figure}[t]
\centering
\includegraphics[width=\linewidth]{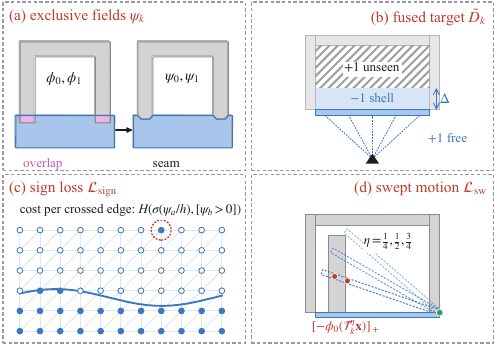}
\caption{\textbf{Optimizing the part fields.} \textbf{(a)}~Raw fields of the frame and the door overlap (left). The exclusive fields give each point to the part that is most inside, so the parts meet at a seam (right). \textbf{(b)}~The target is outside in free space, follows the observed depth at the surface, is inside only within a shell $\Delta$ behind it, and is outside where nothing was seen. \textbf{(c)}~Every sign change on a grid edge costs, so a lone flip without image support is removed. \textbf{(d)}~A door swept along a wrong joint enters the static divider, which the swept loss penalizes.}
\label{fig:fields}
\end{figure}

\subsection{Partitioned Marching Tetrahedra}
\label{sec:meshing}

Meshing one global surface mixes the parts, and remeshing each part separately needs to know where the parts are. We split the problem in two (Fig.~\ref{fig:framework}B,C). The divide step \mD decides which region of space belongs to which part, and the conquer step \mC builds the surface of each part inside its region.

\paragraph{Divide. Articulated tetrahedral partition.}
\mD The divide step takes the depth of both states, the joints, and the labeled triangle model, and returns one cell $\Omega_k$ of tetrahedra per part. We back-project the depth of both states into points. A point seen in state $1$ is assigned the part $k$ of the nearest posed vertex of the triangle model and is mapped back to the canonical frame by $\mathcal{T}_k^{-1}$. Points that fall into the same cell of a voxel grid, with cells $1.5$ times the vertex spacing of the triangle model, are merged into their mean, and a Delaunay tetrahedralization $\mathcal{D}$ of the merged points fills the convex hull of the object with tetrahedra $\tau$ (Fig.~\ref{fig:framework}B). We first decide which tetrahedra are inside the object. A camera ray travels through empty space until it reaches the observed depth, so every tetrahedron crossed by the segment between a camera and its observed point is empty (Fig.~\ref{fig:partition}a). Let $f(\tau)$ count these segments over all views of both states, where the segments of state $1$ are mapped to the canonical frame by the joint of the most likely part of $\tau$ (Fig.~\ref{fig:partition}b). We minimize
\begin{equation}
E_{\mathrm{cut}}(\mathbf{o})=\sum_{\tau\in\mathcal{D}}c_\tau(o_\tau)+\beta\!\!\sum_{(\tau,\tau')\in\mathcal{N}}\!\!A_{\tau\tau'}\,[o_\tau\neq o_{\tau'}]
\label{eq:cut}
\end{equation}
over binary labels $o_\tau\in\{\mathrm{in},\mathrm{out}\}$, where $\mathcal{N}$ holds pairs of tetrahedra sharing a face of area $A_{\tau\tau'}$. The cost $c_\tau(\mathrm{in})$ grows with $f(\tau)$, and $c_\tau(\mathrm{out})$ with the observed surface points just in front of which $\tau$ lies (Sec.~\ref{supp:impl}). The area term prevents noisy votes from punching holes or leaving floating cells. The energy is submodular, so a minimum $s$-$t$ cut gives its exact minimizer~\cite{graphcut,labatut}. Each inside tetrahedron then takes a part label by $\alpha$-expansion with the same face-area term (Sec.~\ref{supp:impl}). A moving part cannot pass through other matter. We pose the tetrahedra of each moving part at $\eta\in\{\tfrac14,\tfrac12,\tfrac34\}$, and an inside tetrahedron of another part that they pass through goes to the moving part if it touches it and to the outside otherwise (Fig.~\ref{fig:partition}c). Small solids detached from a part go to the part they touch (Sec.~\ref{supp:impl}). The result is one cell $\Omega_k$ per part (Fig.~\ref{fig:framework}B). Because the labels partition space, two cells can share a seam but never overlap.

\begin{table*}[t]
\centering
\caption{\textbf{Results on Articulate-100.} Errors per part-count bin and over all 100 objects. All methods receive the number of moving parts and no part masks, and all are scored with the same protocol (Sec.~\ref{supp:eval}). \textbf{Best} and \underline{second best} are marked. CD in mm. Gain is the improvement over the best baseline in each column.}
\label{tab:main}
\setlength{\tabcolsep}{5pt}
\renewcommand{\arraystretch}{0.85}
\resizebox{0.75\linewidth}{!}{
\begin{tabular}{llccccc}
\toprule
 & & 2-3 Parts (54) & 4 Parts (13) & 5 Parts (18) & 6+ Parts (15) & All (100) \\
\midrule
\multirow{3}{*}{Axis Ang ($^\circ$) $\downarrow$}
 & ArtGS~\cite{artgs} & 5.37 & 17.50 & \underline{13.92} & \underline{22.26} & \underline{11.02} \\
 & REArtGS++~\cite{reartgspp} & \textbf{3.92} & \underline{17.47} & 24.69 & 28.43 & 13.10 \\
 & \oc \ArtTet (Ours) & \oc \underline{4.40} & \oc \textbf{8.16} & \oc \textbf{12.62} & \oc \textbf{13.83} & \oc \textbf{7.78} \\
\rowcolor{gray!12}  & Gain & $-$0.48 & +9.31 & +1.30 & +8.43 & +3.24 \\
\midrule
\multirow{3}{*}{Axis Pos (0.1\,m) $\downarrow$}
 & ArtGS & 0.61 & 1.46 & \underline{0.76} & 1.63 & 0.90 \\
 & REArtGS++ & \underline{0.42} & \underline{0.58} & 0.90 & \underline{1.38} & \underline{0.67} \\
 & \oc \ArtTet (Ours) & \oc \textbf{0.21} & \oc \textbf{0.00} & \oc \textbf{0.18} & \oc \textbf{0.23} & \oc \textbf{0.18} \\
\rowcolor{gray!12}  & Gain & +0.21 & +0.58 & +0.58 & +1.15 & +0.49 \\
\midrule
\multirow{3}{*}{Part Motion ($^\circ$/m) $\downarrow$}
 & ArtGS & \underline{5.75} & 10.78 & \underline{8.38} & 14.37 & 8.17 \\
 & REArtGS++ & 6.21 & \underline{6.37} & 9.34 & \underline{12.05} & \underline{7.67} \\
 & \oc \ArtTet (Ours) & \oc \textbf{2.28} & \oc \textbf{0.35} & \oc \textbf{2.41} & \oc \textbf{3.65} & \oc \textbf{2.26} \\
\rowcolor{gray!12}  & Gain & +3.47 & +6.02 & +5.97 & +8.40 & +5.41 \\
\midrule
\multirow{3}{*}{CD-s (mm) $\downarrow$}
 & ArtGS & \underline{19.92} & \underline{26.30} & \underline{23.49} & \underline{28.75} & \underline{22.72} \\
 & REArtGS++ & 24.17 & 32.38 & 32.84 & 47.07 & 30.23 \\
 & \oc \ArtTet (Ours) & \oc \textbf{17.92} & \oc \textbf{21.75} & \oc \textbf{23.00} & \oc \textbf{25.05} & \oc \textbf{20.40} \\
\rowcolor{gray!12}  & Gain & +2.00 & +4.55 & +0.49 & +3.70 & +2.32 \\
\midrule
\multirow{3}{*}{CD-m (mm) $\downarrow$}
 & ArtGS & 75.01 & 116.72 & 168.16 & 173.09 & 111.91 \\
 & REArtGS++ & \underline{74.63} & \underline{107.61} & \underline{165.04} & \underline{166.14} & \underline{108.92} \\
 & \oc \ArtTet (Ours) & \oc \textbf{40.46} & \oc \textbf{42.30} & \oc \textbf{122.52} & \oc \textbf{104.93} & \oc \textbf{65.14} \\
\rowcolor{gray!12}  & Gain & +34.17 & +65.31 & +42.52 & +61.21 & +43.78 \\
\bottomrule
\end{tabular}}
\end{table*}

\paragraph{Conquer. Per-part differentiable marching tetrahedra.}
\mC The partition decides ownership, but its surface is only as smooth as the tetrahedralization. We therefore represent each part by its own signed distance field $\phi_k$, stored at the nodes $\mathbf{g}$ of one shared grid. The grid splits every cube of a regular lattice over the padded bounding box into six tetrahedra along its main diagonal~\cite{kuhn}. This split is conforming, so neighboring tetrahedra agree on their shared faces. The unknowns are the values of every $\phi_k$ at the grid nodes and a color grid. They are initialized by space carving from both states and the partition (Sec.~\ref{supp:impl}), and optimized by gradient descent on the loss of Eq.~\eqref{eq:loss}, through marching tetrahedra and differentiable rendering of both states. The output is one closed mesh per part. To keep the parts apart we define the exclusive field (Fig.~\ref{fig:framework}C and Fig.~\ref{fig:fields}a)
\begin{equation}
\psi_k(\mathbf{g})=\max\Big(\phi_k(\mathbf{g}),\ \phi_k(\mathbf{g})-\min_{j\neq k}\phi_j(\mathbf{g})\Big).
\label{eq:exclusive}
\end{equation}
It is negative exactly where $\phi_k<0$ and $\phi_k<\phi_j$ for every $j\neq k$, so it keeps the interior of part $k$ only where part $k$ is the most inside. Marching tetrahedra~\cite{marchingtets,dmtet} extracts the zero level set of $\psi_k$. On every grid edge $(\mathbf{g}_a,\mathbf{g}_b)$ with $\psi_k(\mathbf{g}_a)<0<\psi_k(\mathbf{g}_b)$ it places the vertex $\mathbf{m}$
\begin{equation}
\mathbf{m}=\mathbf{g}_a+\frac{\psi_k(\mathbf{g}_a)}{\psi_k(\mathbf{g}_a)-\psi_k(\mathbf{g}_b)}\,(\mathbf{g}_b-\mathbf{g}_a),
\label{eq:mt}
\end{equation}
and connects the vertices of each crossed tetrahedron into one or two triangles. The vertex positions are differentiable in $\psi_k$, hence in every $\phi_j$. The construction has two useful properties (proofs in Sec.~\ref{supp:proofs}).

\noindent\emph{Closed part meshes.} The zero level set of a piecewise-linear function on a conforming tetrahedralization has no boundary, so each $\mathcal{M}_k$ is closed whenever $\psi_k$ is positive on the boundary of the grid. The padding and the free-space target below keep it positive there.

\noindent\emph{No overlap.} At every node $\psi_k+\psi_l\ge\max(\phi_k,0)+\max(\phi_l,0)\ge0$ for $k\neq l$, and linear interpolation keeps this inequality inside each tetrahedron, so two part meshes can touch along a seam but cannot interpenetrate at rest.

\noindent\emph{Rendering.} The mesh of part $k$ in state $s$ is $\mathcal{T}_k^{s}(\mathcal{M}_k)$, colored by sampling the color grid trilinearly at each mesh vertex and interpolating across faces. A differentiable rasterizer~\cite{nvdiffrast} renders the image $\hat I_v$, the coverage $\hat\alpha_v$, and the depth $\hat D_v$ of view $v$.

\paragraph{Fused part targets.}
Images constrain the surface weakly where a panel is thin or seen from one side, so each field $\phi_k$ also follows a truncated signed distance target $\bar\phi_k$ fused from the observed depth of both states (Fig.~\ref{fig:fields}b). At a grid node, a view that sees the node in free space votes for outside, and a view that sees it on the surface of part $k$ votes for its signed distance to that surface. A moving part is solid only within a shell of width $\Delta$ behind its observed surface, so the unseen side of a drawer closes just behind its front (Sec.~\ref{supp:impl}).

\paragraph{Objective.}
The conquer step optimizes all fields and the color grid jointly by minimizing
\begin{equation}
\begin{aligned}
\mathcal{L}=\ &\mathcal{L}_{\mathrm{rgb}}+\lambda_{\alpha}\mathcal{L}_{\alpha}+\lambda_{D}\mathcal{L}_{D}+\lambda_{\mathrm{fuse}}\mathcal{L}_{\mathrm{fuse}}\\
&+\lambda_{\mathrm{sign}}\mathcal{L}_{\mathrm{sign}}+\lambda_{\mathrm{lap}}\mathcal{L}_{\mathrm{lap}}+\lambda_{\mathrm{sw}}\mathcal{L}_{\mathrm{sw}} .
\end{aligned}
\label{eq:loss}
\end{equation}
The image term is $\mathcal{L}_{\mathrm{rgb}}=0.8\|\hat I_v-I_v\|_1+0.2\,(1-\mathrm{SSIM}(\hat I_v,I_v))$, and $\mathcal{L}_{\alpha}=\|\hat\alpha_v-A_v\|_1$ matches the silhouette. $\mathcal{L}_{D}$ is an $L_1$ loss on inverse depth over covered pixels inside the silhouette. The fused term pulls each field $\phi_k$, scaled by the truncation $\delta$, toward the fused target $\bar\phi_k$ at the set $\mathcal{O}$ of part and node pairs where the target is defined,
\begin{equation}
\mathcal{L}_{\mathrm{fuse}}=\frac{1}{|\mathcal{O}|}\sum_{(k,\mathbf{g})\in\mathcal{O}}\Big|\operatorname{clip}\!\big(\phi_k(\mathbf{g})/\delta,-1,1\big)-\bar\phi_k(\mathbf{g})\Big| .
\label{eq:lfuse}
\end{equation}
$\mathcal{L}_{\mathrm{sign}}$ is the sign loss of DMTet~\cite{dmtet}, which charges every grid edge the surface crosses, so a surface survives only where the other terms pay for it and floating sign flips vanish (Fig.~\ref{fig:fields}c, Sec.~\ref{supp:impl}). $\mathcal{L}_{\mathrm{lap}}$ penalizes the deviation of $\phi_k$ from its local mean near each surface. The last term encodes collision-free motion (Fig.~\ref{fig:fields}d). Points $\mathbf{x}$ sampled on a moving part are swept along its joint and must stay outside the static body,
\begin{equation}
\mathcal{L}_{\mathrm{sw}}=\frac{1}{3K}\sum_{k=1}^{K}\sum_{\eta\in\{\frac14,\frac12,\frac34\}}\operatorname*{mean}_{\mathbf{x}\in\mathcal{M}_k}\Big[-\phi_0\big(\mathcal{T}_k^{\eta}(\mathbf{x})\big)/h\Big]_+ ,
\label{eq:sweep}
\end{equation}
with $\phi_0$ sampled trilinearly and the grid spacing $h$ (weights in Sec.~\ref{supp:impl}).

\section{Experiments}
\label{sec:exp}

\subsection{Setup}
\label{sec:exp:setup}

\paragraph{Data and metrics.}
We evaluate on Articulate-100~\cite{artmesh}, 100 PartNet-Mobility~\cite{sapien} objects from 16 categories with two to more than six parts, rendered with RGB-D views at two states. For real data, we use the nine SplArt captures~\cite{splart}, which have one moving part each and no sensor depth (Sec.~\ref{supp:real}). Predicted moving parts are paired one to one with the ground truth by the Hungarian algorithm on their Chamfer distance, and the same pairing is used for the joints (Sec.~\ref{supp:eval}). We report the axis angle and axis line distance, the motion error in degrees or meters, and the Chamfer distances CD-s and CD-m in millimeters on the static body and the moving parts, from $10{,}000$ surface samples. For mesh quality we report the shares of open-boundary edges, of manifold edges, and of consistently wound interior edges, averaged over part meshes.

\paragraph{Baselines.}
ArtGS~\cite{artgs} and REArtGS++~\cite{reartgspp} receive, like \ArtTet, the number of moving parts and no part masks. Their meshes are extracted by TSDF fusion, and \ArtTet extracts its meshes with marching tetrahedra (Sec.~\ref{sec:meshing}). Implementation details are in Sec.~\ref{supp:impl}, for the baselines in Sec.~\ref{supp:baselines}.

\par\vspace{0pt plus 5\baselineskip}\penalty-300\vspace{0pt plus -5\baselineskip}
\subsection{Results on Articulate-100}
\label{sec:exp:main}

\begin{figure*}[t]
\centering
\includegraphics[width=\linewidth]{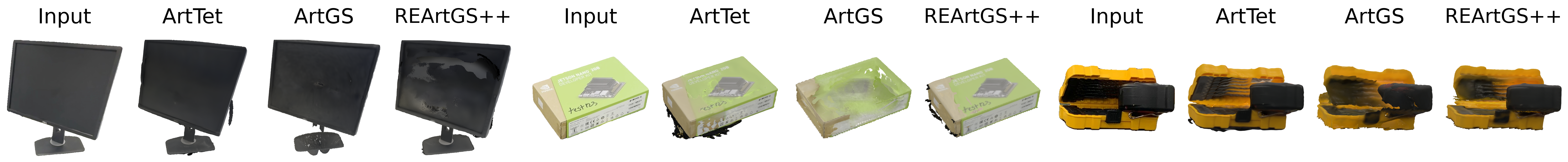}
\caption{\textbf{Real captures from SplArt~\cite{splart}.} Input photo and textured part meshes in the second state. Both states are in Fig.~\ref{fig:supp_splart}.}
\label{fig:real}
\end{figure*}

\begin{figure}[t]
\centering
\includegraphics[width=\linewidth]{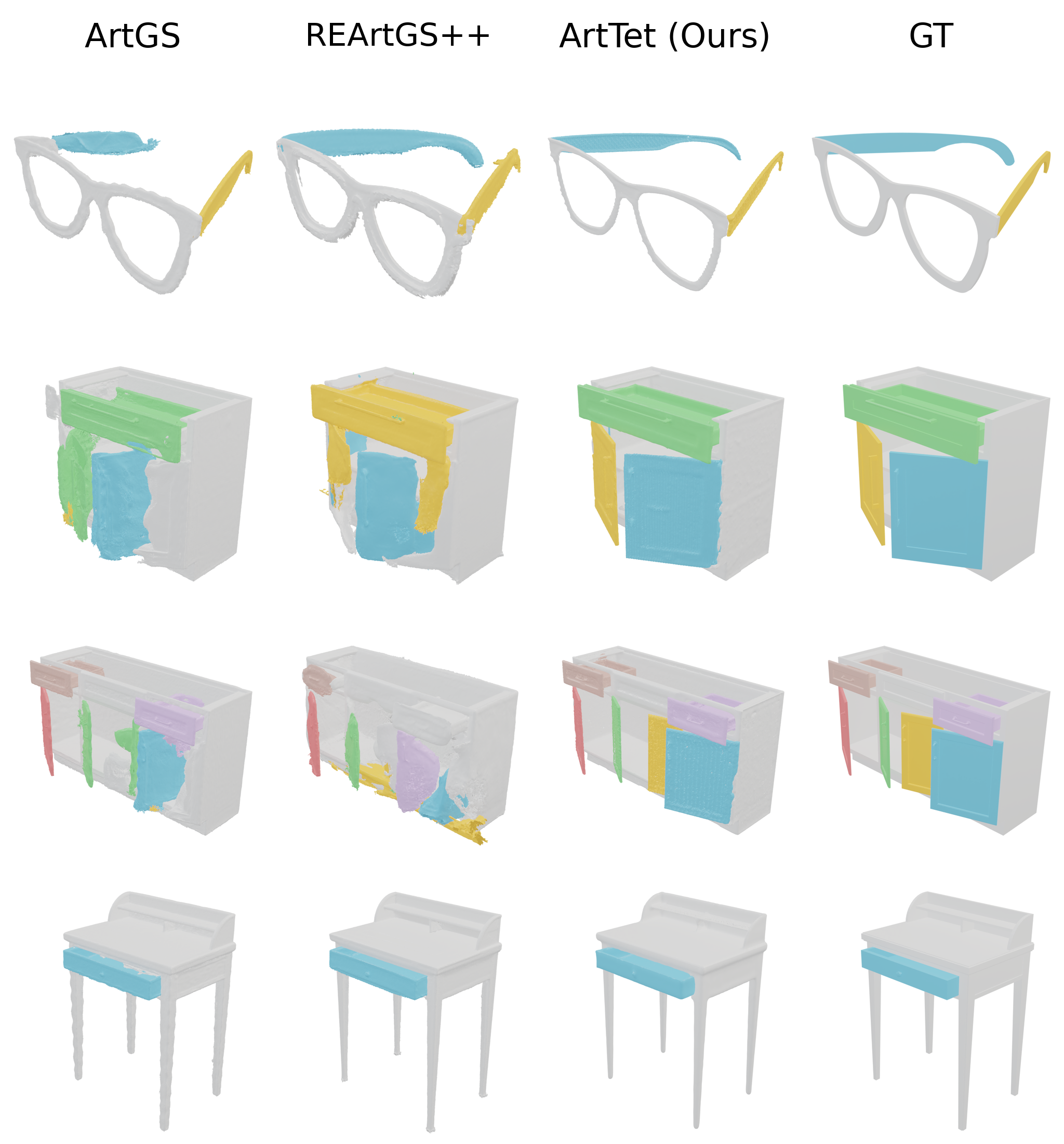}
\caption{\textbf{Qualitative comparison on Articulate-100.} Part meshes in state 1, colored by part; both states are in Fig.~\ref{fig:supp_qual_both}. Gaussian methods are shown after TSDF fusion, which merges neighboring parts and leaves holes and blobs on thin panels. \ArtTet keeps doors and drawers as separate closed meshes that move about the recovered joints.}
\label{fig:qual}
\end{figure}

Table~\ref{tab:main} reports the errors. Over all objects \ArtTet has the lowest axis error, $7.78^\circ$ against $11.02^\circ$ and $13.10^\circ$, the lowest axis position error, $0.18$ against $0.90$ and $0.67$, the lowest motion error, $2.26$ against $8.17$ and $7.67$, and the lowest CD-s and CD-m, $20.40$ and $65.14$ against at best $22.72$ and $108.92$. Our axis position, motion, static, and moving-part errors are the lowest in every bin, and so is our axis error from four parts up. With two or three parts REArtGS++ has the lower axis error, $3.92^\circ$ against $4.40^\circ$. The gaps in axis error and moving-part error are both largest with four parts, $8.16^\circ$ against $17.47^\circ$ and $42.30$ against $107.61$\,mm. Of the $287$ joints, \ArtTet predicts the wrong type for $25$, ArtGS for $30$, and REArtGS++ for $42$.

On the SplArt captures (Fig.~\ref{fig:real}), we compare the predicted joints with hand-labeled ones, computed as in Table~\ref{tab:main} (supplementary Table~\ref{tab:splart_real}). \ArtTet has the lowest axis angle, axis position, and part motion errors, $5.29^\circ$, $0.62$, and $2.71$ against at best $14.58^\circ$, $2.28$, and $13.11$.

\par\vspace{0pt plus 5\baselineskip}\penalty-300\vspace{0pt plus -5\baselineskip}
\subsection{Mesh Quality}
\label{sec:exp:quality}

The TSDF meshes of ArtGS and REArtGS++ are manifold and consistently wound but open, with $2.09\%$ and $2.41\%$ of edges on a boundary, averaged over part meshes. The part meshes of \ArtTet are level sets of fields on a conforming grid (Sec.~\ref{sec:meshing}), so they are closed by construction. Two one-cell export holes are the only boundary edges, and no edge is non-manifold (Sec.~\ref{supp:eval}). Figs.~\ref{fig:teaser}, \ref{fig:qual}, and~\ref{fig:supp_wire} show examples.

\raggedbottom
\begin{table}[t]
\centering
\caption{\textbf{Ablation of dynamic- and geometric-constrained articulation and ownership inference} on Articulate-100, from one shared initialization, before reconstruction.}
\label{tab:abl_physics}
\setlength{\tabcolsep}{3pt}\small
\begin{tabular}{lccc}
\toprule
Variant & Axis ($^\circ$) $\downarrow$ & Pos $\downarrow$ & Motion $\downarrow$ \\
\midrule
Initialization & 9.12 & 0.21 & 2.27 \\
w/o symmetric energy & 8.28 & 0.21 & 2.31 \\
w/o ownership repair & 8.15 & 0.19 & \textbf{2.24} \\
w/o second geometric pass & 8.86 & 0.19 & 2.27 \\
\oc Full refinement & \oc \textbf{7.76} & \oc \textbf{0.19} & \oc 2.26 \\
\bottomrule
\end{tabular}
\end{table}

\begin{table}[t]
\centering
\caption{\textbf{Ablation of the meshing} on Articulate-100 with the same parts and joints. N.C.\ is the normal consistency, and edges are counted after merging vertices at equal positions (Sec.~\ref{supp:eval}). CD in mm. The boundary share $0.00$ is rounded: two one-cell export holes are the only boundary edges (Sec.~\ref{sec:exp:quality}). $^\dagger$Our adaptation of the restricted Delaunay step of MeshSplatting~\cite{meshsplatting}.}
\label{tab:abl_mesh}
\setlength{\tabcolsep}{2.5pt}
\resizebox{\linewidth}{!}{
\begin{tabular}{lccccc}
\toprule
Mesh & CD-s $\downarrow$ & CD-m $\downarrow$ & Bnd.\ (\%) $\downarrow$ & Manif.\ (\%) $\uparrow$ & N.C.\ (\%) $\uparrow$ \\
\midrule
TSDF fusion & 24.91 & \underline{63.33} & 9.12 & \textbf{100.00} & \textbf{100.00} \\
Restricted Delaunay$^\dagger$ & 24.11 & \textbf{60.67} & 6.22 & 50.00 & 33.40 \\
Partition only & \underline{22.85} & 68.01 & \underline{0.11} & 98.75 & \textbf{100.00} \\
\oc Partitioned marching tets & \oc \textbf{20.40} & \oc 65.14 & \oc \textbf{0.00} & \oc \underline{99.99} & \oc \textbf{100.00} \\
\bottomrule
\end{tabular}}
\end{table}

\begin{figure}[t]
\centering
\includegraphics[width=\linewidth]{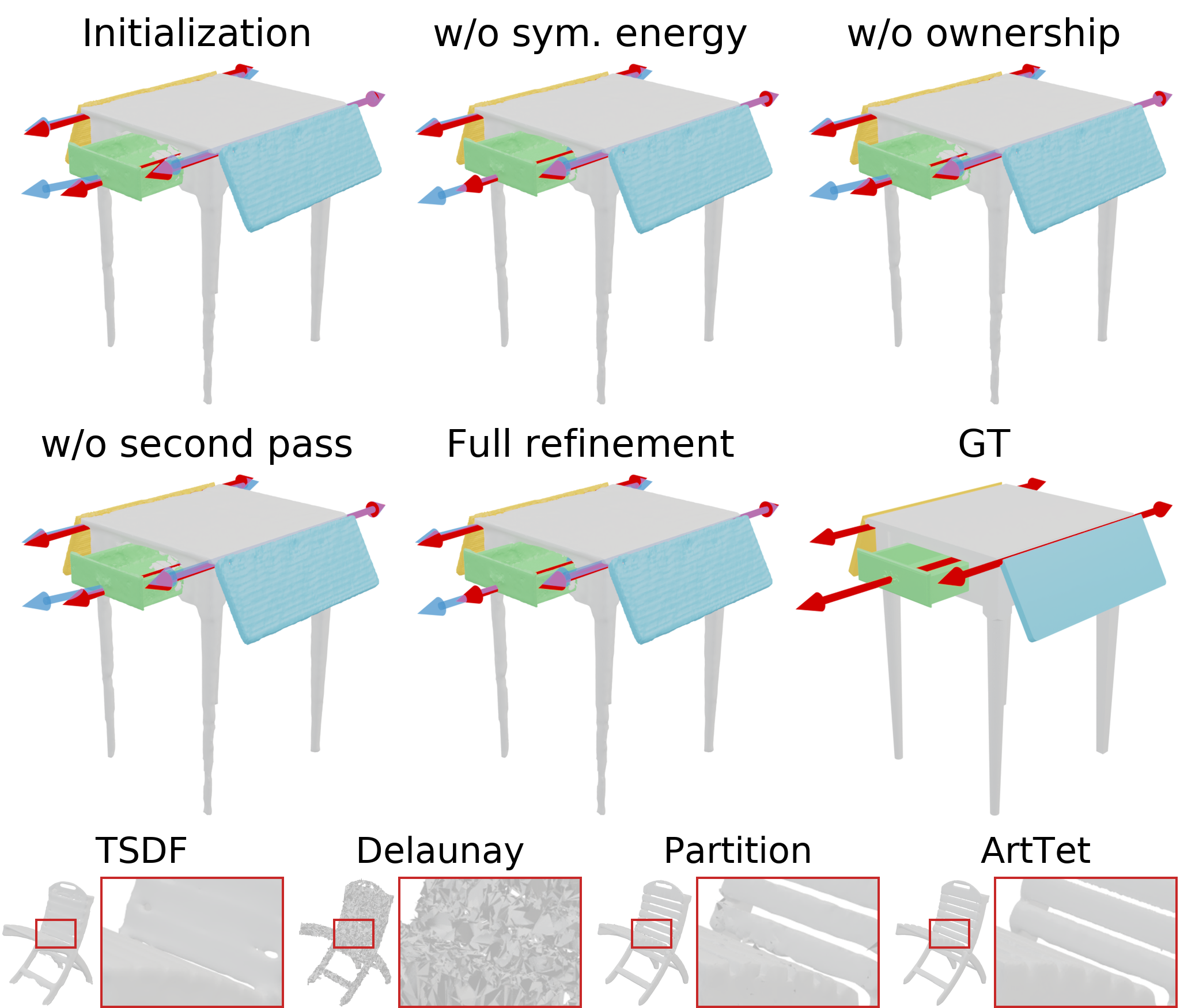}
\caption{\textbf{Ablations.} Top, part labels and joint axes (red) in state 1, with the ground-truth axes in transparent blue for each row of Table~\ref{tab:abl_physics} and the ground truth, colored by part. Bottom, a second object in state 0 meshed by TSDF fusion, restricted Delaunay, the partition alone, and partitioned marching tetrahedra, each with a close-up of the boxed region to its right.}
\label{fig:ablation}
\end{figure}

\par\vspace{0pt plus 5\baselineskip}\penalty-300\vspace{0pt plus -5\baselineskip}
\subsection{Ablations}
\label{sec:exp:ablation}

\paragraph{Articulation and ownership inference.}
From one shared initialization (Table~\ref{tab:abl_physics}), the refinement lowers the axis error from $9.12^\circ$ to $7.76^\circ$. Removing the symmetric energy, the ownership repair, or the second geometric pass raises it to $8.28^\circ$, $8.15^\circ$, or $8.86^\circ$. The refinement lowers the axis error by more than $0.5^\circ$ on 33 objects and raises it by as much on 4. In Fig.~\ref{fig:ablation} (top), the axes of the initialization and of the variants without the ownership repair or the second pass deviate from the ground truth by $2.3^\circ$ to $2.8^\circ$ on average, and those of the full refinement by $1.0^\circ$.

\paragraph{Partitioned marching tetrahedra.}
In Table~\ref{tab:abl_mesh} and Fig.~\ref{fig:ablation}, TSDF fusion leaves $9.12\%$ of edges open. Restricted Delaunay has the lowest CD-m, $60.67$, but half of its edges are non-manifold and only a third of them are consistently oriented. Partitioned marching tetrahedra closes the parts and gives the lowest static error, $20.40$, while its CD-m is $1.8$\,mm above TSDF fusion.

\section{Conclusion}
\label{sec:conclusion}

We presented \ArtTet, which reconstructs an articulated object as one watertight mesh per rigid part with its joints. It divides the object into rigid parts and space into part cells, and then conquers each cell with an exclusive signed distance field. On Articulate-100 it gives the most accurate axis directions, part motions, and part geometry among mask-free methods, and its part meshes are closed up to a negligible share of edges.

\paragraph{Limitations.}
\ArtTet assumes rigid parts. Part ownership comes from the registration, so two neighboring drawers merged there cannot be split later and move as one (Fig.~\ref{fig:supp_fail}). A side of a part that no view observes is closed a fixed shell $\Delta$ behind the observed surface (Fig.~\ref{fig:fields}b), so the hidden back of a thick part comes out too thin. Future work could optimize the part labels together with the fields, so that the images can move part boundaries and split merged parts, and complete unseen sides with a shape prior such as symmetry.

{
    \small
    \bibliographystyle{ieeenat_fullname}
    \bibliography{main}
}

\clearpage
\appendix
\clearpage
\setcounter{section}{0}
\setcounter{figure}{0}
\setcounter{table}{0}
\setcounter{equation}{0}
\renewcommand{\thesection}{\Alph{section}}
\renewcommand{\thefigure}{S\arabic{figure}}
\renewcommand{\thetable}{S\arabic{table}}
\renewcommand{\theequation}{S\arabic{equation}}
\maketitlesupplementary

This supplementary material gives the implementation details of \ArtTet (Sec.~\ref{supp:impl}), the evaluation protocol (Sec.~\ref{supp:eval}), the proofs of the two mesh properties stated in the main paper (Sec.~\ref{supp:proofs}), additional results (Sec.~\ref{supp:results}), further analysis of the results (Sec.~\ref{supp:analysis}), additional baselines (Sec.~\ref{supp:extra_baselines}), and the surface accuracy of the reconstructions (Sec.~\ref{supp:surface}). Equation, section, and table numbers without an S prefix refer to the main paper.

\section{Implementation Details}
\label{supp:impl}

This section lists the settings of \ArtTet. Real captures use the additions of Sec.~\ref{supp:real}.

\subsection{Multibody Registration}
The depth of all views of each state is back-projected and fused into one point cloud per state, subsampled to $60{,}000$ points. Each point's distance to its nearest neighbor in the other state is averaged over its eight nearest neighbors, and points whose averaged distance exceeds the larger of $2.5$ times its median and $2\%$ of its maximum are marked as moving. The label assignment of the multibody initialization (Sec.~\ref{sec:arttet_registration}) is initialized by spectral clustering and $k$-means of the moving points into $K$ groups and by grouping points that move together, and each group gets an initial rigid motion from trimmed least squares~\cite{umeyama} on matched pairs. The alternation of Eq.~\eqref{eq:arttet_procrustes} runs from state $0$ to state $1$ on the moving points. A second registration runs from state $1$ to state $0$; its inverted motions are matched to the forward parts, and each matched pair whose rotations agree within $15^\circ$ is averaged. Points outside the moving set are static. From the tenth iteration a moving point is returned to the static body only when the identity explains it clearly better, with a residual below $0.8$ times that of the best moving part. A part with fewer than $30$ points, or a rotation below $0.5^\circ$ with a translation below $1\%$ of the object diagonal, is re-initialized once from the $5\%$ of moving points that the current motions explain worst, which recovers small parts such as drawers.

\subsection{Joint Typing and Cleanup}
The typing test with the residual of Eq.~\eqref{eq:sy_residual}, measured on the moving points, uses $\kappa=1.3$ and is skipped, keeping the part revolute, when fewer than $20$ mutual pairs exist. Parts of the same joint type whose axes lie within $30^\circ$ of each other then share their residual-weighted mean axis when this does not raise their mean forward residual by more than $2\%$. The cleanup runs two rounds. Each relabels outlier points whose residual exceeds three times the median of their part, re-solves every part's translation with its rotation held fixed, and applies the re-typing with the attachment of Eq.~\eqref{eq:sy_attach} to revolute fits with angles between $2^\circ$ and $\theta_s=13.5^\circ$, with $\kappa'=1.1$ and $\bar a=0.5$ part sizes, comparing forward residuals.

\subsection{Joint and Ownership Refinement}
The refinement of Sec.~\ref{sec:arttet_registration} runs on the registration clouds of $60{,}000$ points. \emph{Ownership repair.} Let $U_k(x)=\min\big(d(T_kx,P^1),\chi\big)/\chi$ be the truncated state-$1$ residual of a point $x$ under the motion of part $k$. A fragment $F$ cut off from part $a$ goes to the touching part $b$ with the lowest $Q(F,b)$ if
\begin{equation}
Q(F,b)+\gamma < Q(F,a),\qquad Q(F,k)=\frac{1}{|F|}\sum_{x\in F}U_k(x),
\label{eq:arttet_transfer}
\end{equation}
with the margin $\gamma>0$. The swept-motion transfer sweeps each moving part along its joint on an occupancy grid, and static material that the part would pass through, but does not touch at rest, goes to the part when the joint carries it onto the state-$1$ observation. \emph{Zero pitch.} A fitted motion $Rx+b$ with rotation axis $u$ equals $R(x-p)+p+(u^\top b)u$ with $p=(I-R)^+(I-uu^\top)b$, a rotation about the pivot $p$ followed by a slide $(u^\top b)u$ along the axis, its pitch. At the end of the refinement each hinge is projected to the pivot form of Eq.~\eqref{eq:arttet_motion} by $b\leftarrow(I-R)p$, which keeps its axis and pivot. The sample spacing $\ell$ is the larger median nearest-neighbor spacing of the two clouds. Each energy minimization runs five rounds of correspondence search per moving part, each followed by three reweighted fits, with pairs trimmed at the $80$th percentile in each direction and the two directions given equal total weight, and parts rotating less than $2^\circ$ are refit as translations. Ownership repair uses a contact graph of radius $2.5$ times the median sample spacing of $P^0$, the margin $\gamma=0.1$ on the truncated residual $U_k$ of Eq.~\eqref{eq:arttet_transfer}, and absorbs static fragments below $2\%$ of the static body. Swept-motion transfer rasterizes the object on a $256^3$ occupancy grid and sweeps each moving part to $\eta\in\{\frac14,\frac12,\frac34,1\}$. The second pass of the geometric constraint is one cleanup round that refits translations and re-types small-angle hinges with the symmetric residual $r_k$, without outlier relabeling or axis consensus, and zero pitch is accepted when $r_k$ rises by at most $5\%$. Every step uses only the two observed clouds.

In the initialization, each clustering method runs with two seeds, and half of the groupings match neighbors in position and color. Within each grouping the joints are typed and revolute fits are polished by point-to-plane ICP, and the grouping with the lowest sum of forward and coverage residuals is kept. During the initialization both terms of $r_k$ are measured on the moving samples only. The energy uses the Huber penalty $\rho(r)=r^2/2$ for $|r|\le1$ and $|r|-\frac12$ otherwise, with weight $1/\max(1,|r|/\ell)$, and the confidence $w$ is the number of neighbors within $1.5\%$ of the object diagonal divided by its median and clipped to $[0,1]$; each part is normalized by its total weight. Parts with fewer than $30$ samples in $\mathcal{X}_k$ or $\mathcal{Y}_k$ are not refined. A hinge that fails the gate keeps its pitch, and the swept transfer, which runs before the projection, follows the full screw path of Eq.~\eqref{eq:arttet_motion}. About a third of the hinges that rotate by $5^\circ$ or more fail the gate and keep both $\theta_k$ and $d_k$ of Eq.~\eqref{eq:arttet_motion} instead of the revolute case $d_k=0$; for half of them the slide along the axis is below $1.1\%$ of the object diagonal. The reconstruction and CD-m use this screw motion, while the axis, axis position, and motion errors read only its rotation, as for the free rigid motions of REArtGS++. In the swept transfer, samples within two grid cells of a differently labeled sample at rest are exempt, and a static sample reached by several parts goes to the one that reaches it most often. The transfer gate is $U_k(x)\le 2\operatorname*{med}_{z_j=k}U_k(x_j)+0.25$, and $\chi$ is four times the median residual of the current labels.

\subsection{Triangle Field and Part Labels}
The canonical triangle model follows MeshSplatting~\cite{meshsplatting} and is trained for $30{,}000$ iterations with the joints fixed. Views of state $1$ are rendered after posing each vertex by its part's joint, with the gradient kept, so both states supervise geometry and appearance. The losses are the photometric term $0.8\,L_1+0.2\,(1-\mathrm{SSIM})$, a depth term whose weight decays exponentially over training, a normal-consistency term between rendered and surface normals, and the opacity terms of MeshSplatting, and a vertex depth-consistency term from iteration $12{,}000$. Part logits are passed through a softmax whose temperature decreases from $5$ to $0.5$, with a straight-through hard assignment for posing, and are anchored to their current hard labels, reset after each densification or relabel, by $\mathcal{L}_\pi$ of Eq.~\eqref{eq:tri} with $\lambda_\pi=0.01$. During training, pruning keeps all faces of any part that would otherwise be left with fewer than $32$.

The part-label rules of Sec.~\ref{sec:triangle} use the following settings. Rigid-body connectivity first splits each moving part at a radius of six median point spacings. A detached fragment stays with its part only if its state-$1$ residual under its own joint, normalized by four times the median residual and capped at one, is lower by at least $0.3$ than under the receiving part. A second pass makes every part, including the static body, one piece at $2.5$ spacings, where a detached fragment joins a touching part only if that part fits it better by $0.1$, and static fragments below $2\%$ of the static body are always absorbed. The swept-motion transfer poses each moving part at $\eta\in\{\frac14,\frac12,\frac34,1\}$ on a $256^3$ voxel grid, exempts material in contact with the part at rest, and transfers an obstacle to the part only when the joint carries it onto the state-$1$ observation. The first two rules are applied before training and once more at iteration $15{,}000$, after which appearance re-fits for the remaining iterations. Each vertex created by densification between iterations $500$ and $10{,}000$ takes the label shared by at least $90\%$ of its neighbors on its own tangent plane and otherwise keeps its parent's label, so thin parts lying on larger ones are not absorbed.

\subsection{Articulated Tetrahedral Partition}
The partition is built from the observed depth of both states. Points seen in state $1$ are mapped to the canonical frame through the joint of the part they show, and points are averaged on a voxel grid whose cell is $1.5$ times the vertex spacing. The unary cost $c_\tau(\mathrm{in})$ of Eq.~\eqref{eq:cut} grows with $f(\tau)$ and is raised when the centroid of $\tau$ projects onto the background in a view of either state, and $c_\tau(\mathrm{out})$ grows with the number of samples just behind observed surface points that fall in $\tau$, so that it is high for tetrahedra lying just behind the observed surface along the camera ray, with a constant addition when $\tau$ was never seen as free in either state. The minimum cut of Eq.~\eqref{eq:cut} is computed with a max-flow solver on the adjacency graph of the tetrahedra. Part labels are computed by $\alpha$-expansion with a face-area Potts term, using the four vertex labels and the free space and observed surface of state $1$ under each joint. Sweep exclusion tests the tetrahedra of each moving part at $\eta\in\{\frac14,\frac12,\frac34\}$. Each part keeps every connected solid that holds at least $5\%$ of the volume of its largest solid, and smaller solids go to the part with which they share the most faces, or to the outside. Tetrahedra behind outside faces farther than $2\%$ of the bounding-box diagonal from every observed point of their part are removed, ring by ring, for at most $40$ rings and $60\%$ of a part's tetrahedra. Closed shells below $1\%$ of a part's area are then dropped, and each part surface receives ten rounds of Taubin smoothing. The partition provides ownership only. Its surface is not part of the final output.

\subsection{Partitioned Marching Tetrahedra}
\begin{table}[t]
\centering
\caption{\textbf{Hyperparameters of partitioned marching tetrahedra.} $h$ is the grid spacing.}
\label{tab:supp_hparams}
\setlength{\tabcolsep}{3pt}\small
\begin{tabular}{@{}p{0.4\linewidth}p{0.57\linewidth}@{}}
\toprule
Parameter & Value \\
\midrule
Grid & $192^3$ cubes, six tetrahedra per cube \\
Bounding box & soup vertices (0.2 to 99.8 percentile), padded by $6\%$ \\
Color grid & $256^3$ \\
Iterations & $3{,}000$, one training view per iteration \\
Learning rate & $0.03\,h$ for $\phi_k$, $0.01$ for color, decayed $\times0.1$ \\
Render supersampling & $2\times$ \\
Active band & $4h$ around each surface, refreshed every 25 iterations \\
Carving & margin $1.5h$, at least 2 views \\
Truncation $\delta$ & $3h$ \\
Ownership ratio $\nu$ & $0.5$, when the owning part has at least 3 surface observations \\
Shell width $\Delta$ & $4h$ \\
$\lambda_\alpha,\lambda_D,\lambda_{\mathrm{fuse}},\lambda_{\mathrm{sign}},\lambda_{\mathrm{sw}}$ & $1,\ 0.05,\ 3,\ 0.02,\ 1$ \\
$\lambda_{\mathrm{lap}}$ & $0.2$, rising linearly to $2$ after $30\%$ of iterations \\
Image gradient to geometry & $1$, fading linearly to $0$ after $30\%$ of iterations \\
Export & pieces below $0.5\%$ of a part's area removed \\
\bottomrule
\end{tabular}
\end{table}
Table~\ref{tab:supp_hparams} lists the settings. The grid stores one value of $\phi_k$ per part at each of its $193^3$ nodes. The six tetrahedra of a cube follow the six monotone paths from one corner to the opposite corner, which makes the split conforming. The color grid is sampled at mesh vertices and interpolated across faces by the rasterizer. Each iteration draws one training view from either state, renders the posed part meshes with nvdiffrast~\cite{nvdiffrast}, and minimizes the weighted sum of Eq.~\eqref{eq:loss}. The active band restricts marching tetrahedra to tetrahedra near a current surface, which keeps an iteration fast at $192^3$. The fade of the image gradient is implemented by rendering colors from detached vertex positions blended with the attached ones. On every grid edge $(\mathbf{g}_a,\mathbf{g}_b)$ crossed by the surface of part $k$, $\mathcal{L}_{\mathrm{sign}}$ applies the binary cross-entropy $H\big(\sigma(\psi_k(\mathbf{g}_a)/h),[\psi_k(\mathbf{g}_b)>0]\big)$ and its swap, with the sigmoid $\sigma$ and the grid spacing $h$. \emph{Fused part targets.} The partition of the divide step is rasterized into every view to give a part label map $L_v$. A node $\mathbf{g}$ is posed by $\mathcal{T}_k^{s}$ into each view $v$ of state $s$ and projected to pixel $\mathbf{u}$ at camera depth $z$, with $e_v=D_v(\mathbf{u})-z$. The view sees the node as free when $\mathbf{u}$ lies outside the silhouette or $e_v>\delta$, and as a surface point of part $k$ when $|e_v|\le\delta$ and $L_v(\mathbf{u})=k$. With $n^{\mathrm{free}}_k$ free views and the set $\mathcal{V}_k$ of surface views, the target is
\begin{equation}
\bar\phi_k(\mathbf{g})=\frac{n^{\mathrm{free}}_k(\mathbf{g})+\sum_{v\in\mathcal{V}_k(\mathbf{g})}\operatorname{clip}\!\big(e_v/\delta,-1,1\big)}{n^{\mathrm{free}}_k(\mathbf{g})+|\mathcal{V}_k(\mathbf{g})|},
\label{eq:fuse}
\end{equation}
defined at nodes with at least two observations. Nodes farther than $\Delta$ from every surface observation of part $k$ get $\bar\phi_k=1$ when they are otherwise unobserved. A node observed as the surface of several parts is outside for part $k$, $\bar\phi_k=1$, when $|\mathcal{V}_k|<\nu\max_j|\mathcal{V}_j|$. After $30\%$ of the iterations the image gradient to the geometry fades and $\lambda_{\mathrm{lap}}$ rises (Table~\ref{tab:supp_hparams}), so the shape is then set by the silhouette, the depth, the fused targets, and smoothness, and the color keeps fitting the images.

\subsection{Real Captures}
\label{supp:real}
\begin{figure*}[t]
\centering
\includegraphics[width=\linewidth]{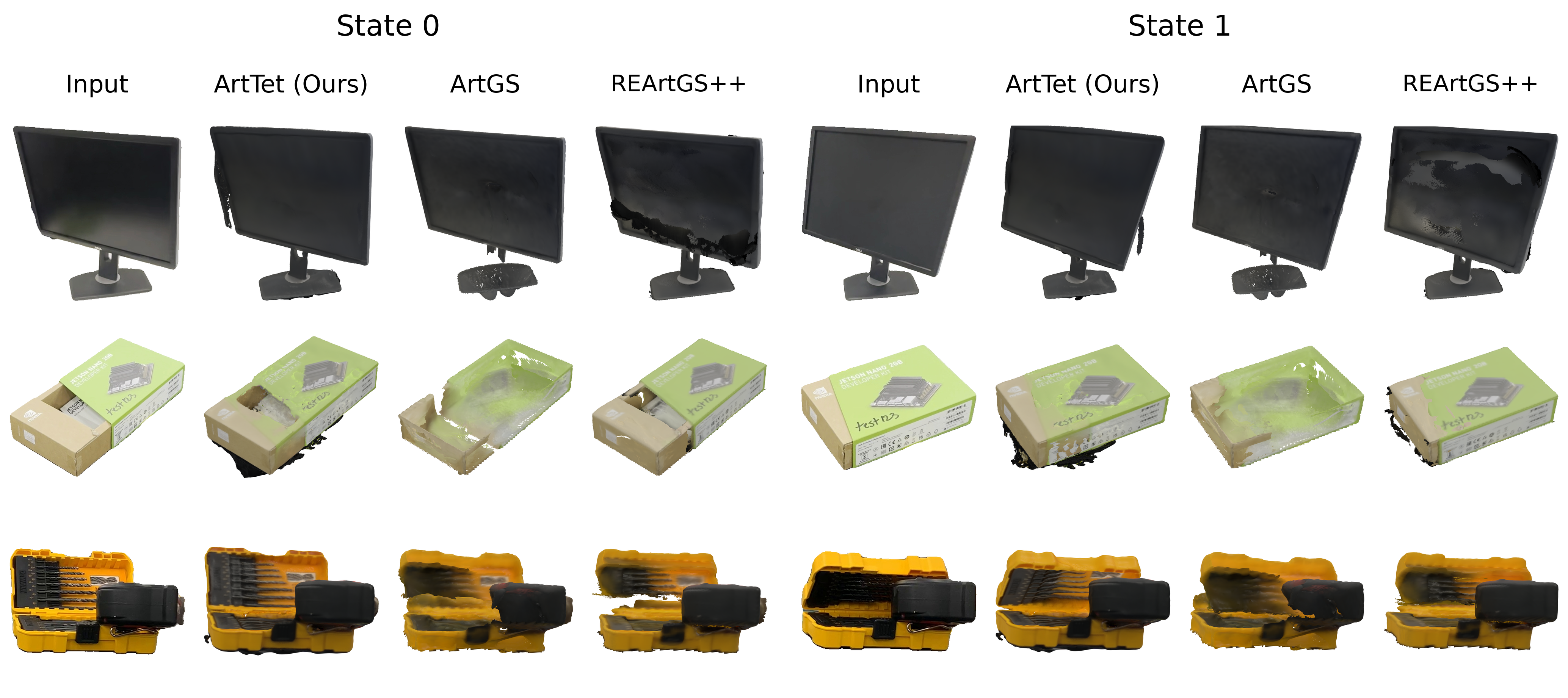}
\caption{\textbf{Real captures from SplArt in both states.} Input photos and the textured part meshes of \ArtTet, ArtGS, and REArtGS++. Gaussian methods are shown after TSDF fusion.}
\label{fig:supp_splart}
\end{figure*}

SplArt~\cite{splart} provides posed images and object masks but no depth. We prepare their input depth from the images and masks. For each state we carve the visual hull from the masks, scale the monocular depth~\cite{depthanything} of every view to the front surface of the hull, fuse it, and normalize the scene with it. We then fit a single-state triangle model to the images and masks of each state, initialized on the surface of the visual hull and trained without any depth loss, and render its depth into every view, keeping the fused depth on the few mask pixels the model does not cover. This depth agrees across views by construction, and the rest of the pipeline is unchanged (Fig.~\ref{fig:supp_splart}). Since real masks are less precise than rendered ones, the carving margin, the truncation $\delta$, and the shell width are widened to $8h$, and a node is carved only when at least half of the views that see it see through it.

\subsection{Hardware}
Registration and its refinement run on the CPU and take about $6$ to $10$ minutes per object. The triangle field, the partition, and the marching-tetrahedra fields then take about $25$ minutes per object on one NVIDIA A800 GPU.

\subsection{Baselines}
\label{supp:baselines}
ArtGS~\cite{artgs} and REArtGS++~\cite{reartgspp} are run with their released code and default settings on the same training views. Each receives $K+1$ part slots for the $K$ moving parts and no part masks, and initializes the part centers from its own coarse Gaussian stage. ArtGS then trains its articulation stage for $20{,}000$ iterations with its joint-type prior, and REArtGS++ for $30{,}000$ iterations. Their meshes are extracted by TSDF fusion.

\section{Evaluation Details}
\label{supp:eval}
We use the metrics of ArtGS~\cite{artgs} with one pairing for all metrics and apply them to every method in Table~\ref{tab:main}. Each state is evaluated with the moving parts posed by the recovered joints, using $10{,}000$ area-weighted surface samples on prediction and ground truth. Predicted moving parts are paired one to one with the ground-truth moving parts by the Hungarian algorithm on their Chamfer distance averaged over the two states, and CD-m is the mean distance of the pairs. The same pairing is used for the joints. Each predicted joint is read under its own type. ArtGS and \ArtTet output a type for each part. REArtGS++ outputs a rigid motion for each part, which we read as revolute if it rotates by $5^\circ$ or more and as prismatic otherwise, the cutoff that \ArtTet uses for its own types. The direction of a slide is the displacement of the center of its part, which does not depend on the coordinate frame. Axis Ang compares the predicted joint axis with the ground truth, Axis Pos the distance between revolute axis lines, with prismatic joints counted as zero, and Part Motion the difference of the part motions, the geodesic angle between the predicted and ground-truth rotations for revolute joints and the distance between the translations for prismatic joints. A joint of the wrong type keeps its axis angle error and takes the failure penalty below for axis position, when the ground truth is revolute, and for motion. A ground-truth part without a partner, or whose predicted axis position is degenerate, takes the failure penalty of $90^\circ$ axis error, $1$\,m axis position error, $90^\circ$ or $1$\,m motion error, and $1{,}000$\,mm Chamfer distance. No ground-truth part of Articulate-100 is left without a partner or with a degenerate axis by \ArtTet. CD-s is computed on the static body and averaged over the two states.

The mesh-quality metrics are computed per part mesh and averaged. An edge is a boundary edge when it belongs to one face and a manifold edge when it belongs to at most two. Normal consistency is the share of interior edges whose two faces traverse the edge in opposite directions, so that their orientations agree. The meshes of ArtGS and REArtGS++ for this comparison are extracted by TSDF fusion.

\paragraph{Why \ArtTet is not at $100\%$.} The level sets of Sec.~\ref{sec:meshing} are closed when no node value is exactly zero (Sec.~\ref{supp:proofs}). Over all $100$ objects and $2.7\times10^8$ edges, the exported part meshes have no non-manifold edge. The $99.99\%$ is the mean over part meshes; pooled over all edges, the metric finds $0.0042\%$ non-manifold, and these appear only because the metric first merges vertices at equal positions. Where a field is nearly zero at a grid node, interpolation in floating point places the vertices of several crossing edges exactly at that node. Merging them turns each triangle with two vertices at that node into a degenerate triangle whose two remaining edges coincide with each other and with an edge of a neighboring triangle, so the metric counts that edge as shared by more than two faces. The only boundary edges are $15$ edges in Table\_33930 and Toilet\_102621, each a single closed loop of about one grid cell in the static body, away from the boundary of the grid. They come from the export, not from the level sets: marching tetrahedra runs only on tetrahedra within the active band of Table~\ref{tab:supp_hparams}, and a few crossed tetrahedra fell outside it.

\begin{figure}[t]
\centering
\includegraphics[width=\linewidth]{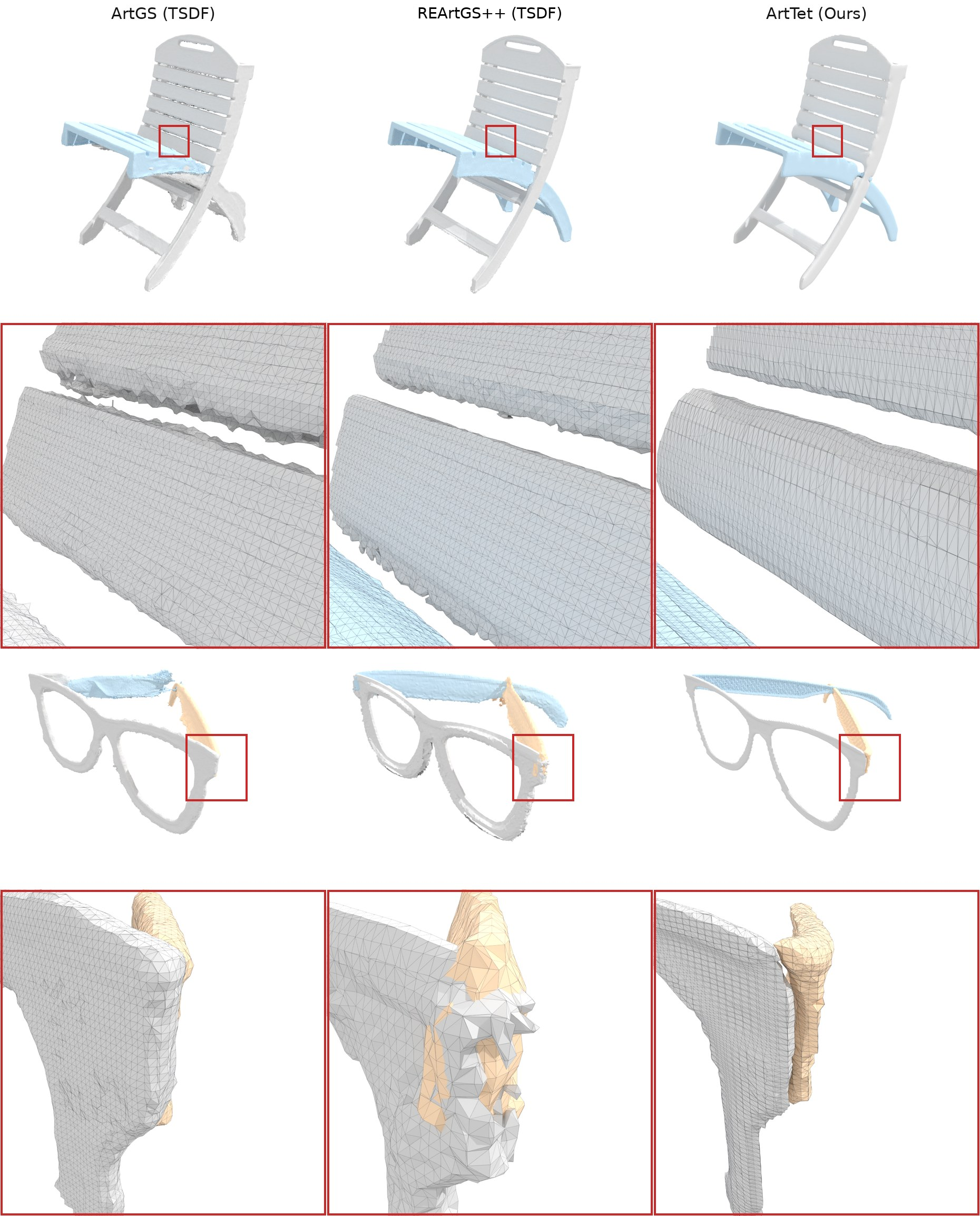}
\caption{\textbf{Mesh structure.} A folding chair and a pair of glasses in state 0, colored by part (top of each pair), and a wireframe close-up of the boxed region (bottom): the gap between two seat slats and the hinge of a temple. TSDF fusion gives ArtGS and REArtGS++ ragged slat edges and a temple fused into the frame. In \ArtTet the temple is a separate closed mesh beside the frame.}
\label{fig:supp_wire}
\end{figure}

\paragraph{Mesh structure.} Fig.~\ref{fig:supp_wire} shows the triangles of the three methods. \ArtTet's part meshes have $1.78$M faces and $0.89$M vertices per object on average, against $1.76$M and $0.89$M for the TSDF meshes of ArtGS and $1.57$M and $0.80$M for REArtGS++, from one shared grid of $193^3$ nodes.

\section{Proofs of the Mesh Properties}
\label{supp:proofs}

Let $\tilde\psi_k$ denote the piecewise-linear interpolation of the node values $\psi_k(\mathbf{g})$ over the tetrahedral grid. Marching tetrahedra outputs, inside each tetrahedron, the zero set of the linear function $\tilde\psi_k$ restricted to it. With one negative node this is one triangle, and with two negative nodes it is a planar quadrilateral split into two triangles, so the output is exactly the zero set of $\tilde\psi_k$ wherever it crosses a tetrahedron.

\paragraph{Closed meshes.}
Assume no node value is exactly zero, which holds almost surely and can be enforced with an arbitrarily small offset. Every triangle edge produced by marching tetrahedra either lies in a face of the tetrahedron that produced it or is the diagonal that splits a quadrilateral cut, which is shared by the two triangles of that cut inside the same tetrahedron. For an edge in a face, because the split is conforming, that face is shared by exactly one other tetrahedron unless it lies on the boundary of the grid. Both tetrahedra see the same three node values on the shared face, so both produce the same segment on it, and every interior triangle edge is shared by exactly two triangles. If $\psi_k>0$ on the boundary of the grid, no triangle edge lies on the boundary, so the mesh has no boundary edge.

\paragraph{No overlap.}
Write $m_k=\min_{j\neq k}\phi_j$, so that $\psi_k=\max(\phi_k,\phi_k-m_k)=\phi_k+\max(0,-m_k)$. For $k\neq l$, the minimum $m_k$ runs over a set that contains $l$, so $m_k\le\phi_l$ and $\max(0,-m_k)\ge\max(0,-\phi_l)$. In the same way $\max(0,-m_l)\ge\max(0,-\phi_k)$. Adding the two expansions gives, at every node,
\begin{equation}
\begin{aligned}
\psi_k+\psi_l\ &\ge\ \big(\phi_k+\max(0,-\phi_k)\big)+\big(\phi_l+\max(0,-\phi_l)\big)\\
&=\ \max(\phi_k,0)+\max(\phi_l,0)\ \ge\ 0 .
\end{aligned}
\end{equation}
Linear interpolation preserves the sign of a sum of node values, so $\tilde\psi_k+\tilde\psi_l\ge0$ at every point of the grid. The two interpolated fields therefore cannot both be negative at the same point, and the interiors of the part meshes are disjoint. They can meet only where both fields are zero, which is a shared seam.

\section{Additional Results}
\label{supp:results}

\begin{figure}[!htb]
\centering
\includegraphics[width=\linewidth]{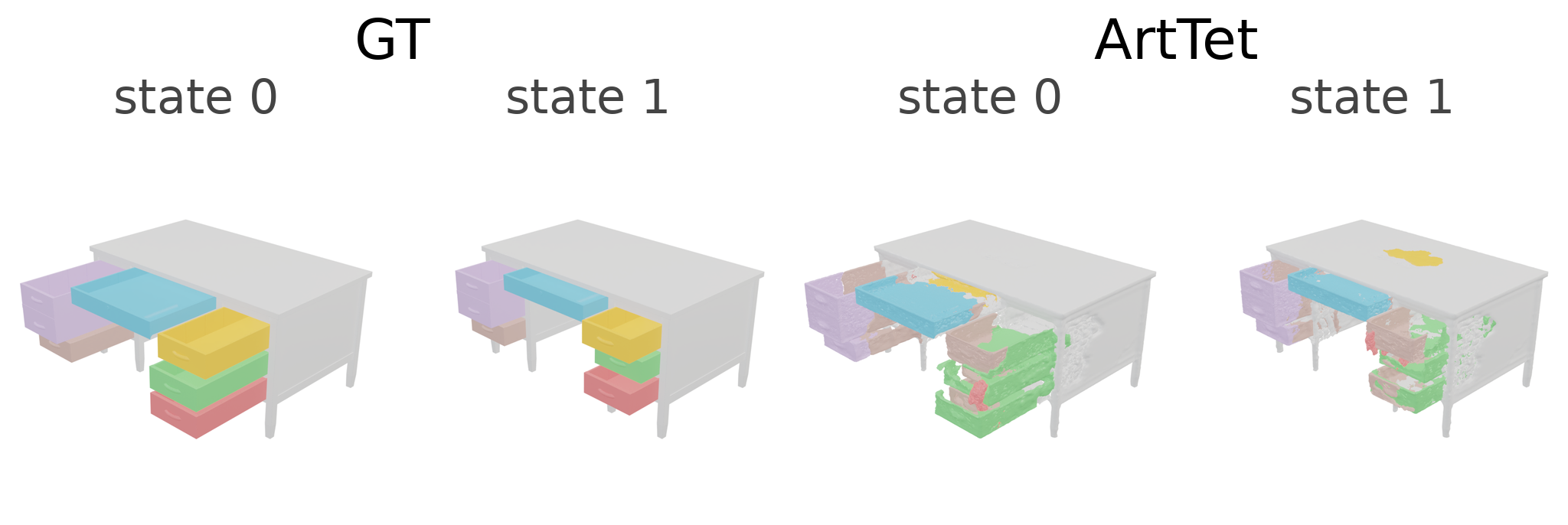}
\caption{\textbf{Failure case.} Registration merges neighboring drawers of the right stack into one part (green). Every later stage keeps them together, so they move as one, and stray fragments of a drawer end up on the desk top.}
\label{fig:supp_fail}
\end{figure}

\begin{figure*}[t]
\centering
\includegraphics[width=\linewidth]{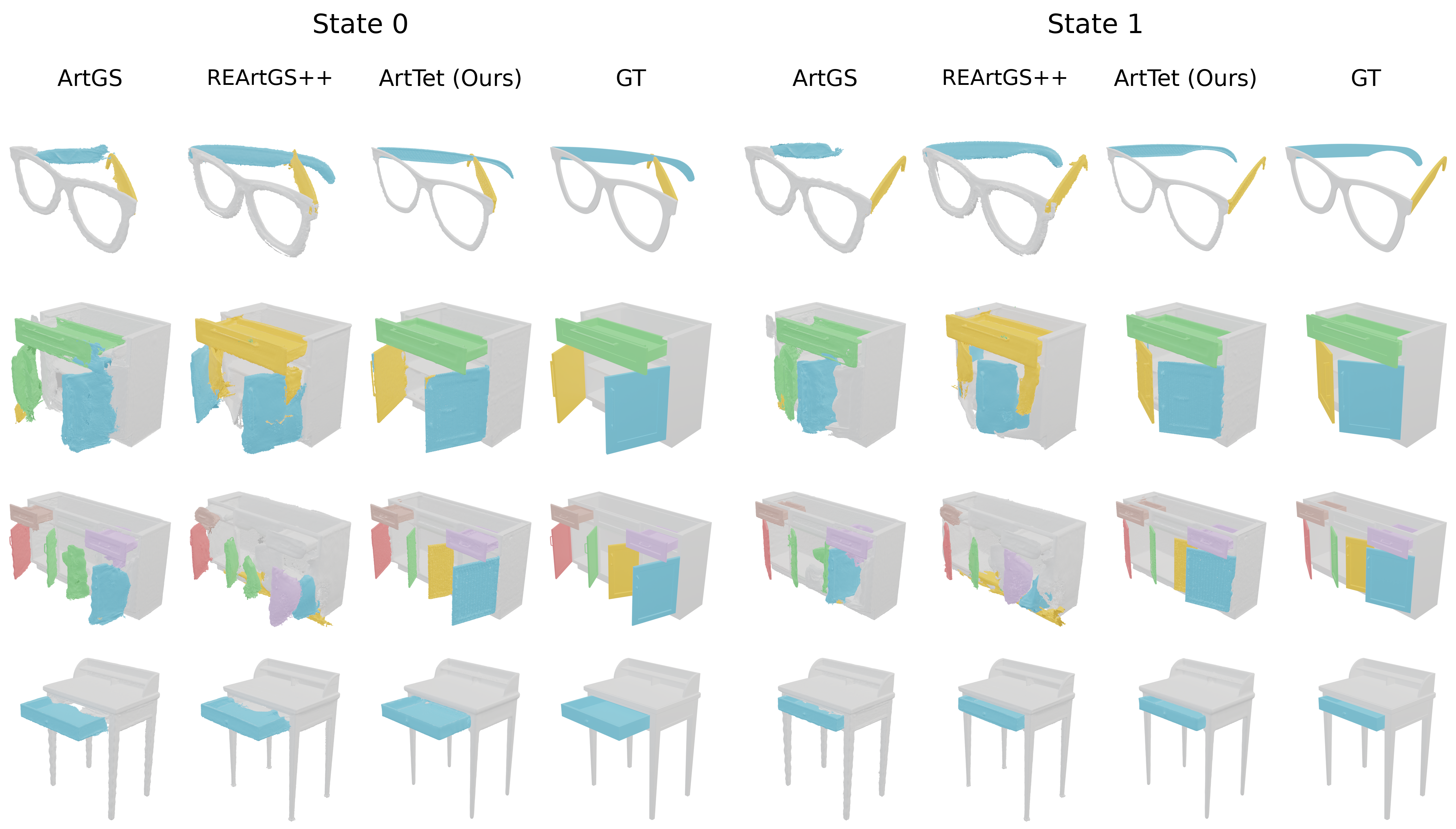}
\caption{\textbf{Qualitative comparison on Articulate-100 in both states.} The objects of Fig.~\ref{fig:qual}, colored by part. Gaussian methods are shown after TSDF fusion.}
\label{fig:supp_qual_both}
\end{figure*}

\begin{figure*}[p]
\centering
\includegraphics[width=\linewidth,height=0.86\textheight,keepaspectratio]{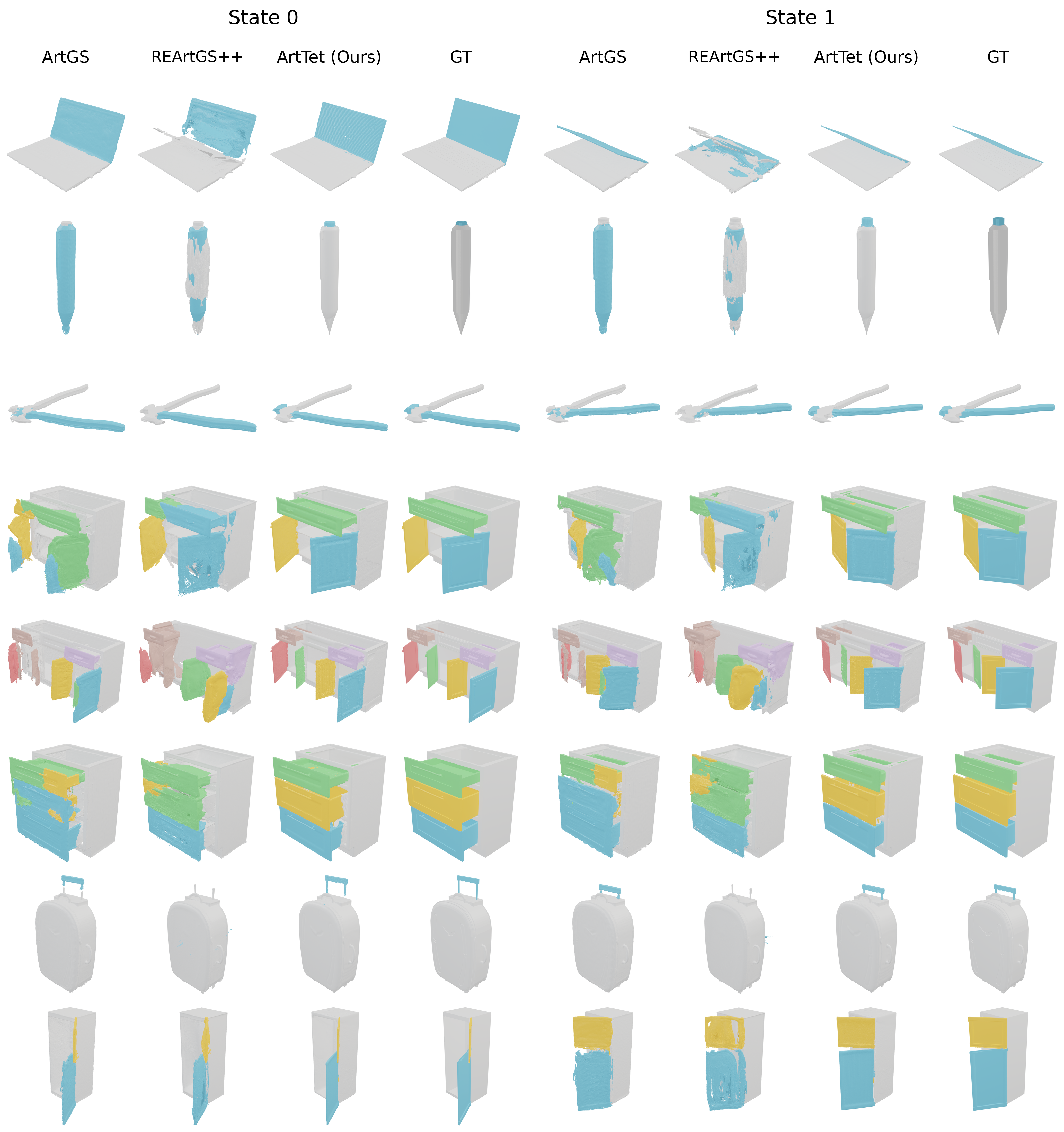}
\caption{\textbf{Additional qualitative results on Articulate-100.} ArtGS, REArtGS++, \ArtTet, and GT in both states, colored by part. Gaussian methods are shown after TSDF fusion.}
\label{fig:supp_qual}
\end{figure*}

\subsection{Joint Accuracy on Real Captures}
\label{supp:real_quant}
\begin{table}[t]
\centering
\caption{\textbf{Quantitative results on the SplArt real captures}, against joints labeled by hand, computed as in Table~\ref{tab:main} (Sec.~\ref{supp:eval}): each prediction is read under its own joint type, and the errors are averaged over the nine captures. SplArt has no metric scale, so lengths are in units $u$ of its reconstruction frame. \textbf{Best} and \underline{second best} are marked.}
\label{tab:splart_real}
\setlength{\tabcolsep}{3pt}\small
\begin{tabular}{lccc}
\toprule
Method & Axis ($^\circ$) $\downarrow$ & Pos ($0.1u$) $\downarrow$ & Motion ($^\circ$/$u$) $\downarrow$ \\
\midrule
ArtGS~\cite{artgs} & \underline{14.58} & \underline{2.28} & \underline{13.11} \\
REArtGS++~\cite{reartgspp} & 17.16 & 2.83 & 18.82 \\
\oc \ArtTet (Ours) & \oc \textbf{5.29} & \oc \textbf{0.62} & \oc \textbf{2.71} \\
\bottomrule
\end{tabular}
\end{table}

SplArt provides no ground-truth joints, so we triangulated the joint of each of the nine captures from points marked by hand in several input photos, which give the axis direction, the hinge line of a revolute joint, and the motion between the two states. All methods are mapped into the frame of these labels; \ArtTet works in a normalized frame, which we map back by a similarity fitted to the camera centers. Table~\ref{tab:splart_real} reports the errors, computed as for Articulate-100. \ArtTet has the lowest error in all three measures. Its mean axis error is $5.29^\circ$ against $14.58^\circ$ and $17.16^\circ$, it places the hinge lines most precisely, $0.62$ against $2.28$ and $2.83$, and its part motion error is $2.71$ against $13.11$ and $18.82$. The pen, whose cap slides only a short distance, is the hardest capture, with an axis error of $18.8^\circ$.

Fig.~\ref{fig:supp_fail} shows the most common failure. Part ownership comes from the registration, so two neighboring drawers that the registration merges stay merged, and the meshes of both drawers move as one part. Fig.~\ref{fig:supp_qual} shows more objects from Articulate-100.

\section{Further Analysis}
\label{supp:analysis}
This section separates the geometry of \ArtTet from its joint accuracy and ablates the reconstruction end to end on all $100$ objects of Articulate-100.

\subsection{Geometry and Joint Accuracy}
\label{supp:geom_joint}
\begin{table}[t]
\centering
\caption{\textbf{Geometry separated from joint accuracy} on Articulate-100. Left: state $0$ only, where no joint is applied, with moving parts paired on their state-$0$ Chamfer distance. Right: CD-m of Table~\ref{tab:main}, split by the mean axis error of \ArtTet on each object. CD in mm. \textbf{Best} is marked.}
\label{tab:supp_geom}
\setlength{\tabcolsep}{3pt}\small
\resizebox{\linewidth}{!}{
\begin{tabular}{lcccc}
\toprule
 & \multicolumn{2}{c}{State $0$ (100)} & \multicolumn{2}{c}{CD-m by \ArtTet axis error} \\
\cmidrule(lr){2-3}\cmidrule(lr){4-5}
Method & CD-s $\downarrow$ & CD-m $\downarrow$ & $\le15^\circ$ (81) & $>15^\circ$ (19) \\
\midrule
ArtGS~\cite{artgs} & 22.61 & 113.83 & 89.18 & 208.85 \\
REArtGS++~\cite{reartgspp} & 30.12 & 111.34 & 92.63 & 178.35 \\
\oc \ArtTet (Ours) & \oc \textbf{20.35} & \oc \textbf{65.32} & \oc \textbf{49.00} & \oc \textbf{133.96} \\
\bottomrule
\end{tabular}}
\end{table}

The Chamfer distances of Table~\ref{tab:main} pose the moving parts by the recovered joints, so a wrong joint also raises CD-m. Table~\ref{tab:supp_geom} separates the two in two ways. In state $0$ no joint is applied, and predicted moving parts are paired with the ground-truth parts by the Hungarian algorithm on their state-$0$ Chamfer distance, with $1000$\,mm for an unpaired ground-truth part as in Sec.~\ref{supp:eval}. There, \ArtTet has a CD-m of $65.32$\,mm against $113.83$ for ArtGS and $111.34$ for REArtGS++, at least $41\%$ lower, and its state-$0$ errors are close to its two-state errors in Table~\ref{tab:main} ($20.35$ and $65.32$ against $20.40$ and $65.14$\,mm). The right half splits the objects by the mean axis error of \ArtTet. Joint error and moving-part error are correlated (Spearman $0.58$ over the $100$ objects), but \ArtTet has the lowest CD-m both on the $81$ objects with an axis error of at most $15^\circ$ and on the $19$ objects above it. The lower moving-part error therefore does not come from more accurate joints alone.

\subsection{End-to-End Ablation}
\label{supp:abl_e2e}
\begin{table}[t]
\centering
\caption{\textbf{End-to-end ablation} on all 100 objects of Articulate-100. Each row reruns the reconstruction with one component removed, and the full row is \ArtTet as in Table~\ref{tab:main}. Errors are computed as in Table~\ref{tab:main}, with parts paired on the final meshes, so the axis error of the same joints can change with the meshes; for the same reason the initial joints score $9.09^\circ$ here and $9.12^\circ$ in Table~\ref{tab:abl_physics}, which pairs parts before reconstruction. CD in mm. \textbf{Best} is marked.}
\label{tab:supp_abl_e2e}
\setlength{\tabcolsep}{3pt}\small
\begin{tabular}{lccc}
\toprule
Variant & Axis ($^\circ$) $\downarrow$ & CD-s $\downarrow$ & CD-m $\downarrow$ \\
\midrule
w/o exclusive fields & \textbf{7.78} & 20.50 & 66.29 \\
w/o swept loss & \textbf{7.78} & \textbf{20.31} & 65.66 \\
w/o partition labels & 7.82 & 20.87 & 65.43 \\
w/o joint refinement & 9.09 & 20.66 & 67.38 \\
\oc Full & \oc \textbf{7.78} & \oc 20.40 & \oc \textbf{65.14} \\
\bottomrule
\end{tabular}
\end{table}

\begin{table*}[t]
\centering
\caption{\textbf{Additional baselines on Articulate-100.} DTA and Part2GS next to the methods of Table~\ref{tab:main}, whose rows are repeated from there. All methods receive the number of moving parts and no part masks, and all are scored with the protocol of Sec.~\ref{supp:eval}. DTA and Part2GS run their released code, Part2GS with the changes of Sec.~\ref{supp:extra_baselines}. \textbf{Best} and \underline{second best} are marked. CD in mm.}
\label{tab:supp_extra_baselines}
\setlength{\tabcolsep}{5pt}
\renewcommand{\arraystretch}{0.85}
\resizebox{0.75\linewidth}{!}{
\begin{tabular}{llccccc}
\toprule
 & & 2-3 Parts (54) & 4 Parts (13) & 5 Parts (18) & 6+ Parts (15) & All (100) \\
\midrule
\multirow{5}{*}{Axis Ang ($^\circ$) $\downarrow$}
 & ArtGS~\cite{artgs} & 5.37 & 17.50 & \underline{13.92} & \underline{22.26} & \underline{11.02} \\
 & REArtGS++~\cite{reartgspp} & \textbf{3.92} & \underline{17.47} & 24.69 & 28.43 & 13.10 \\
 & DTA~\cite{digitaltwinart} & 6.05 & 28.29 & 21.91 & 32.78 & 15.80 \\
 & Part2GS~\cite{part2gs} & 10.00 & 20.90 & 24.12 & 28.84 & 16.78 \\
 & \oc \ArtTet (Ours) & \oc \underline{4.40} & \oc \textbf{8.16} & \oc \textbf{12.62} & \oc \textbf{13.83} & \oc \textbf{7.78} \\
\midrule
\multirow{5}{*}{Axis Pos (0.1\,m) $\downarrow$}
 & ArtGS & 0.61 & 1.46 & \underline{0.76} & 1.63 & 0.90 \\
 & REArtGS++ & \underline{0.42} & \underline{0.58} & 0.90 & \underline{1.38} & \underline{0.67} \\
 & DTA & 0.61 & 1.19 & 1.98 & 3.49 & 1.37 \\
 & Part2GS & 1.32 & 2.06 & 1.21 & 2.31 & 1.54 \\
 & \oc \ArtTet (Ours) & \oc \textbf{0.21} & \oc \textbf{0.00} & \oc \textbf{0.18} & \oc \textbf{0.23} & \oc \textbf{0.18} \\
\midrule
\multirow{5}{*}{Part Motion ($^\circ$/m) $\downarrow$}
 & ArtGS & \underline{5.75} & 10.78 & 8.38 & 14.37 & 8.17 \\
 & REArtGS++ & 6.21 & \underline{6.37} & 9.34 & \underline{12.05} & \underline{7.67} \\
 & DTA & 7.69 & 11.47 & \underline{6.84} & 18.02 & 9.58 \\
 & Part2GS & 9.89 & 14.11 & 10.01 & 19.78 & 11.94 \\
 & \oc \ArtTet (Ours) & \oc \textbf{2.28} & \oc \textbf{0.35} & \oc \textbf{2.41} & \oc \textbf{3.65} & \oc \textbf{2.26} \\
\midrule
\multirow{5}{*}{CD-s (mm) $\downarrow$}
 & ArtGS & 19.92 & 26.30 & 23.49 & 28.75 & 22.72 \\
 & REArtGS++ & 24.17 & 32.38 & 32.84 & 47.07 & 30.23 \\
 & DTA & \underline{18.38} & \textbf{21.52} & \textbf{21.37} & \textbf{22.20} & \textbf{19.90} \\
 & Part2GS & 20.22 & 25.94 & 23.97 & 28.58 & 22.89 \\
 & \oc \ArtTet (Ours) & \oc \textbf{17.92} & \oc \underline{21.75} & \oc \underline{23.00} & \oc \underline{25.05} & \oc \underline{20.40} \\
\midrule
\multirow{5}{*}{CD-m (mm) $\downarrow$}
 & ArtGS & 75.01 & 116.72 & 168.16 & 173.09 & 111.91 \\
 & REArtGS++ & 74.63 & \underline{107.61} & \underline{165.04} & \underline{166.14} & \underline{108.92} \\
 & DTA & \underline{69.35} & 188.33 & 167.97 & 222.25 & 125.50 \\
 & Part2GS & 94.75 & 208.53 & 202.06 & 172.95 & 140.59 \\
 & \oc \ArtTet (Ours) & \oc \textbf{40.46} & \oc \textbf{42.30} & \oc \textbf{122.52} & \oc \textbf{104.93} & \oc \textbf{65.14} \\
\bottomrule
\end{tabular}}
\end{table*}

Table~\ref{tab:supp_abl_e2e} removes one component at a time and reruns the reconstruction on all $100$ objects. Unlike Table~\ref{tab:abl_mesh}, which exchanges the mesher for the same parts and joints, these variants change the method itself. \emph{w/o exclusive fields} trains independent fields $\phi_k$ in place of Eq.~\eqref{eq:exclusive}, \emph{w/o swept loss} sets $\lambda_{\mathrm{sw}}=0$, \emph{w/o partition labels} takes part ownership from the labeled triangle model instead of the tetrahedral partition, and \emph{w/o joint refinement} runs the whole reconstruction from the initial joints of Table~\ref{tab:abl_physics}. The first three variants retrain only the fields, from the triangle models and partitions of Table~\ref{tab:main}, while \emph{w/o joint refinement} refits the triangle model, the partition, and the fields with the initial joints. Training is not deterministic, and rerunning the complete method this way changes the mean CD-m by $0.21$\,mm. Removing the joint refinement raises the axis error by $1.31^\circ$ and CD-m by $2.24$\,mm, and removing exclusivity raises CD-m by $1.15$\,mm. Without the partition labels, CD-s rises by $0.47$\,mm and CD-m by $0.29$\,mm. Without the swept loss, CD-m rises by $0.52$\,mm and CD-s falls by $0.09$\,mm. The gains of the partition labels and the swept loss in CD-m are close to the variation between runs. The swept loss and exclusivity target overlap between parts rather than distance to the surface, so small Chamfer gains are expected.

\section{Additional Baselines}
\label{supp:extra_baselines}
\begin{figure*}[p]
\centering
\includegraphics[width=\linewidth,height=0.86\textheight,keepaspectratio]{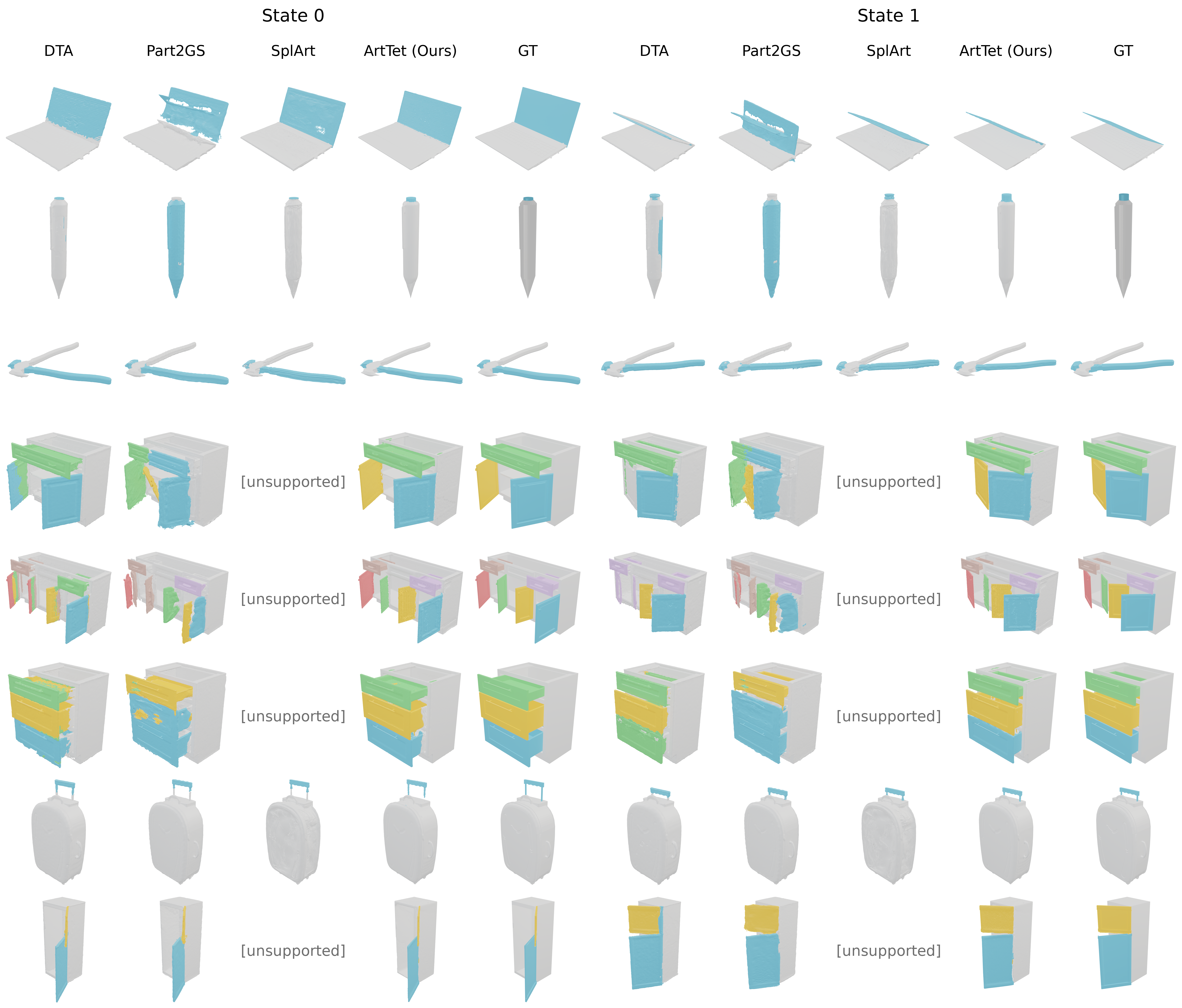}
\caption{\textbf{Additional baselines on Articulate-100.} DTA, Part2GS, SplArt, \ArtTet, and GT on the objects of Fig.~\ref{fig:supp_qual} in both states, colored by part. Part2GS is shown after TSDF fusion. SplArt models a single moving part and is marked unsupported on objects with two or more moving parts.}
\label{fig:supp_qual_extra}
\end{figure*}

We further compare against DTA~\cite{digitaltwinart} and Part2GS~\cite{part2gs} on all $100$ objects of Articulate-100 (Table~\ref{tab:supp_extra_baselines}). As for ArtGS and REArtGS++ (Sec.~\ref{supp:baselines}), each method runs its released code and configuration on the same training views, receives the number of moving parts and no part masks, and is scored with the pairing, joint reading, and failure penalties of Sec.~\ref{supp:eval}. \ArtTet has the lowest axis, axis position, motion, and moving-part errors over all $100$ objects. Its CD-m of $65.14$\,mm compares with $125.50$\,mm for DTA and $140.59$\,mm for Part2GS, and its axis error of $7.78^\circ$ with $15.80^\circ$ and $16.78^\circ$ for DTA and Part2GS. DTA, which reconstructs a signed distance field from the RGB-D images, has the lowest static-body error, with a CD-s of $19.90$\,mm against $20.40$\,mm for \ArtTet, while its moving parts are less accurate, especially on objects with four or more parts. DTA reports experiments with up to three parts. The Chamfer comparisons of the main paper therefore hold against the baselines of Table~\ref{tab:main}; with DTA added, \ArtTet keeps the lowest moving-part and joint errors, and DTA has a slightly lower static-body error. SplArt~\cite{splart} models a single moving part, so we show it only in Fig.~\ref{fig:supp_qual_extra}, which compares the three methods on the objects of Fig.~\ref{fig:supp_qual}.

\textbf{Part2GS implementation.} The released Part2GS code evaluates four losses over all pairs of Gaussians at every iteration (repulsion, neighborhood part consistency, contact, and vector-field alignment), which exceeds $80$\,GB of GPU memory once densification reaches a few $10^4$ Gaussians, also on the authors' own scenes. Following the definition of the part loss in the Part2GS paper over the Gaussians of the current batch, we evaluate these four losses on a random batch of $8192$ Gaussians per iteration, with nearest neighbors still searched over all Gaussians. The released code uses ground-truth joints for the initial part assignment and for checkpoint selection. We replace both and use the clustering-based part assignment included in the same code and the final checkpoint, so that no ground-truth joint information enters training or model selection. We further fix two crashes, in which the per-Gaussian part data is not extended during densification and a visualization indexes at most ten parts, and evaluate the deformation of each state once per iteration. One run takes $22$ to $40$ minutes on one A800 GPU. These changes will be made publicly available with our code.

\section{Surface Accuracy}
\label{supp:surface}
\begin{table}[t]
\centering
\caption{\textbf{Surface accuracy} on Articulate-100, over all $100$ objects. The whole object in state $0$, where no joint is applied, is compared with the ground truth using $60{,}000$ area-weighted surface samples on each. F-score at a threshold of $1\%$ of the ground-truth bounding-box diagonal, and the mean angle between the normal at each predicted sample and the normal at its nearest ground-truth sample, ignoring orientation. The meshes are those of Table~\ref{tab:main}. \textbf{Best} is marked.}
\label{tab:supp_surface}
\setlength{\tabcolsep}{4pt}\small
\begin{tabular}{lcc}
\toprule
Method & F-score $\uparrow$ & Normal error ($^\circ$) $\downarrow$ \\
\midrule
ArtGS~\cite{artgs} & 0.856 & 22.4 \\
REArtGS++~\cite{reartgspp} & 0.743 & 27.4 \\
\oc \ArtTet (Ours) & \oc \textbf{0.910} & \oc \textbf{20.0} \\
\bottomrule
\end{tabular}
\end{table}

Chamfer distance averages over the surface, so we also report the F-score, the share of surface within a fixed distance of the other surface, and the agreement of surface normals (Table~\ref{tab:supp_surface}). Both are computed on the whole object in state $0$, where no joint is applied, so they measure the reconstructed surface alone. \ArtTet has an F-score of $0.910$ against $0.856$ for ArtGS and $0.743$ for REArtGS++, and a normal error of $20.0^\circ$ against $22.4^\circ$ and $27.4^\circ$.

\end{document}